\pdfoutput=1
\documentclass[acmsmall, review=false, screen]{acmart}
\usepackage{verbatim}

\usepackage{footmisc}

\usepackage{times}
\usepackage{epsfig,subfigure,overpic}

\usepackage{amssymb}
\usepackage{multirow}
\usepackage{enumitem}
\usepackage{comment}
\usepackage{color}
\usepackage{url}
\usepackage{threeparttable}

\usepackage{subfigure} 
\usepackage{balance}

\usepackage[colorinlistoftodos]{todonotes}

\usepackage{tablefootnote}

\usepackage{makecell}

\makeatletter
\newcommand\notsotiny{\@setfontsize\notsotiny{6.31415}{7.1828}}
\makeatother

\newcommand*{\origrightarrow}{}

\renewcommand*{\textrightarrow}{\fontfamily{cmr}\selectfont\origrightarrow}

\AtBeginDocument{%
  \providecommand\BibTeX{{%
    \normalfont B\kern-0.5em{\scshape i\kern-0.25em b}\kern-0.8em\TeX}}}

\begin{document}

\title{Fashion Meets Computer Vision: A Survey}


\author{Wen-Huang Cheng}
\affiliation{%
  \institution{National Chiao Tung University and
National Chung Hsing University}
}
\email{whcheng@nctu.edu.tw}

\author{Sijie Song}
\affiliation{%
  \institution{Peking University}
}
\email{ssj940920@pku.edu.cn}

\author{Chieh-Yun Chen}
\affiliation{%
  \institution{National Chiao Tung University}
}
\email{sky4568520.ep05@nctu.edu.tw}

\author{Shintami Chusnul Hidayati}
\affiliation{%
  \institution{Institut Teknologi Sepuluh Nopember}
}
\email{shintami@its.ac.id}

\author{Jiaying Liu}
\authornote{Corresponding author}
\affiliation{%
  \institution{Peking University}
}
\email{liujiaying@pku.edu.cn}



\renewcommand{\shortauthors}{W.-H. Cheng \textit{et al.}}

\begin{abstract}
Fashion is the way we present ourselves to the world and has become one of the world's largest industries. Fashion, mainly conveyed by vision, has thus attracted much attention from computer vision researchers in recent years. Given the rapid development, this paper provides a comprehensive survey of more than 200 major fashion-related works covering four main aspects for enabling intelligent fashion: (1) Fashion detection includes landmark detection, fashion parsing, and item retrieval, (2) Fashion analysis contains attribute recognition, style learning, and popularity prediction, (3) Fashion synthesis involves style transfer, pose transformation, and physical simulation, and (4) Fashion recommendation comprises fashion compatibility, outfit matching, and hairstyle suggestion. For each task, the benchmark datasets and the evaluation protocols are summarized. Furthermore, we highlight promising directions for future research.

\end{abstract}




\maketitle

\section{Introduction}
\label{sec:intro}

Fashion is how we present ourselves to the
world. The way we dress and makeup defines our unique style and distinguishes us from other people. Fashion in modern society has become an indispensable part of who I am. Unsurprisingly, the global fashion apparel market alone has surpassed 3 trillion US dollars today, and accounts for nearly 2 percent of the world's Gross Domestic Product (GDP)\footnote{\url{https://fashionunited.com/global-fashion-industry-statistics/}}. Specifically, revenue in the Fashion segment amounts to over US \$718 billion in 2020 and is expected to present an annual growth of 8.4\%\footnote{\url{https://www.statista.com/outlook/244/100/fashion/worldwide}}.

As the revolution of computer vision with artificial intelligence (AI) is underway, AI is starting to hit the magnanimous field of fashion, whereby reshaping our fashion life with a wide range of application innovations from electronic retailing, personalized stylist, to the fashion design process. In this paper, we term the computer-vision-enabled fashion technology as intelligent fashion. Technically, intelligent fashion is a challenging task because, unlike generic objects, fashion items suffer from significant variations in style and design, and, most importantly, the long-standing semantic gap between computable low-level features 
and high-level semantic concepts that they encode is huge.

There are few previous works~\cite{liu:mmmag2014, Song2018fashion} related to short fashion surveys. In 2014, Liu~\emph{et al.}~\cite{liu:mmmag2014} presented an initial literature survey focused on intelligent fashion analysis with facial beauty and clothing analysis, which introduced the representative works published during 2006--2013. However, thanks to the rapid development of computer vision, there are far more than these two domains within intelligent fashion, e.g., style transfer, physical simulation, fashion prediction. There have been a lot of related works needed to be updated. In 2018, Song and Mei~\cite{Song2018fashion} introduced the progress in fashion research with multimedia, which categorized the fashion tasks into three aspects: low-level pixel computation, mid-level fashion understanding, and high-level fashion analysis. Low-level pixel computation aims to generate pixel-level labels on the image, such as human segmentation, landmark detection, and human pose estimation. Mid-level fashion understanding aims to recognize fashion images, such as fashion items and fashion styles. High-level fashion analysis includes recommendation, fashion synthesis, and fashion trend prediction. However, there is still a lack of a systematic and comprehensive survey to paint the whole picture of intelligent fashion so as to summarize and classify state-of-the-art methods, discuss datasets and evaluation metrics, and provide insights for future research directions.

\begin{figure}[t]
  \centering
  \includegraphics[width=\linewidth]{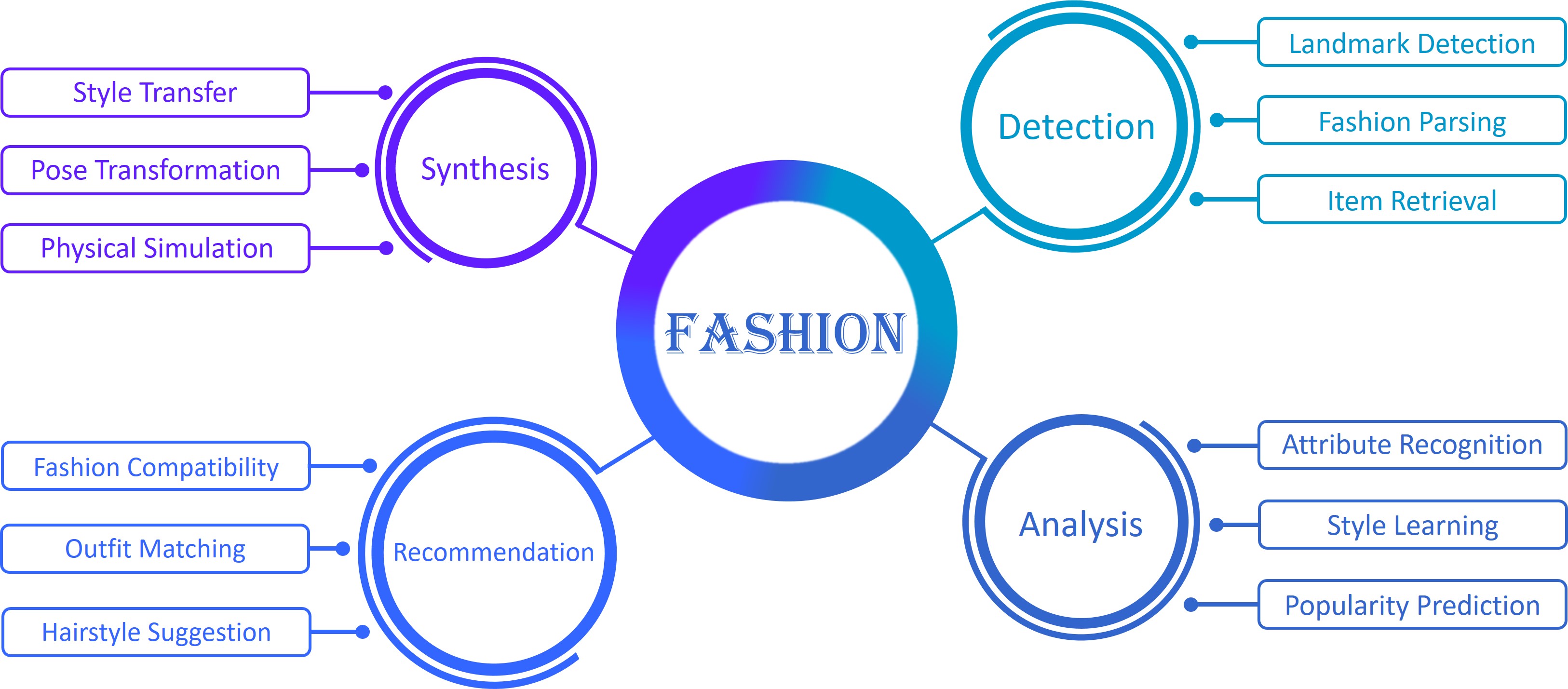}
  \caption{Scope of the intelligent fashion research topics covered in this survey paper.}
  \setlength{\belowcaptionskip}{-80pt}
  \label{fig:overview}
\end{figure}

Current studies on intelligent fashion covers the research topics not only to detect what fashion items are presented in an image but also analyze the items, synthesize creative new ones, and finally provide personalized recommendations. Thus, in this paper, we organize the research topics accordingly, as categorized in Fig.~\ref{fig:overview}, which includes fashion image detection, analysis, synthesis, and recommendation. In addition, we also give an overview of main applications in the fashion domain, showing the power of intelligent fashion in the fashion industry. Overall, the contributions of our work can be summarized as follows:
\begin{itemize}
  \item We provide a comprehensive survey of the current state-of-the-art research progress in the fashion domain and categorize fashion research topics into four main categories: detection, analysis, synthesis, and recommendation.
  \item For each category in the intelligent fashion research, we provide an in-depth and organized review of the most significant methods and their contributions. Also, we summarize the benchmark datasets as well as the links to the corresponding online portals.
  \item We gather evaluation metrics for different problems and also give performance comparisons for different methods.
  \item We list possible future directions that would help upcoming advances and inspire the research community.
\end{itemize}

This survey is organized in the following sections. Sec.~\ref{sec:fffashion_detection} reviews the fashion detection tasks including landmark detection, fashion parsing, and item retrieval. Sec.~\ref{sec:fffashion_analysis} illustrates the works for fashion analysis containing attribute recognition, style learning, and popularity prediction. Sec.~\ref{sec:fffashion_synthesis} provides an overview of fashion synthesis tasks comprising style transfer, human pose transformation, and physical texture simulation. Sec.~\ref{sec:fffashion_recommendation} talks about works of fashion recommendation involving fashion compatibility, outfit matching, and hairstyle suggestion. Besides, Sec.~\ref{sec:application} demonstrates selected  applications and future work. Last but not least, concluding remarks are given in Sec.~\ref{sec:conclusion}.






\section{Fashion Detection}
\label{sec:fffashion_detection}

Fashion detection is a widely discussed technology since most fashion works need detection first. Take virtual try-on as an example~\cite{fashionon}. It needs to early detect the human body part of the input image for knowing where the clothing region is and then synthesize the clothing there. Therefore, detection is the basis for most extended works. In this section, we mainly focus on fashion detection tasks, which are split into three aspects: landmark detection, fashion parsing, and item retrieval. For each aspect, state-of-the-art methods, the benchmark datasets, and the performance comparison are rearranged.


\subsection{Landmark Detection}
\label{subsec:landmark_detection}

Fashion landmark detection 
aims to predict the positions of functional keypoints defined on the clothes, such as the corners of the neckline, hemline, and cuff. These landmarks not only indicate the functional regions of clothes, but also implicitly capture their bounding boxes, making the design, pattern, and category of the clothes can be better distinguished. Indeed, features extracted from these landmarks greatly facilitate fashion image analysis.

\begin{figure}[t!]
  \centering
  \includegraphics[width=\linewidth]{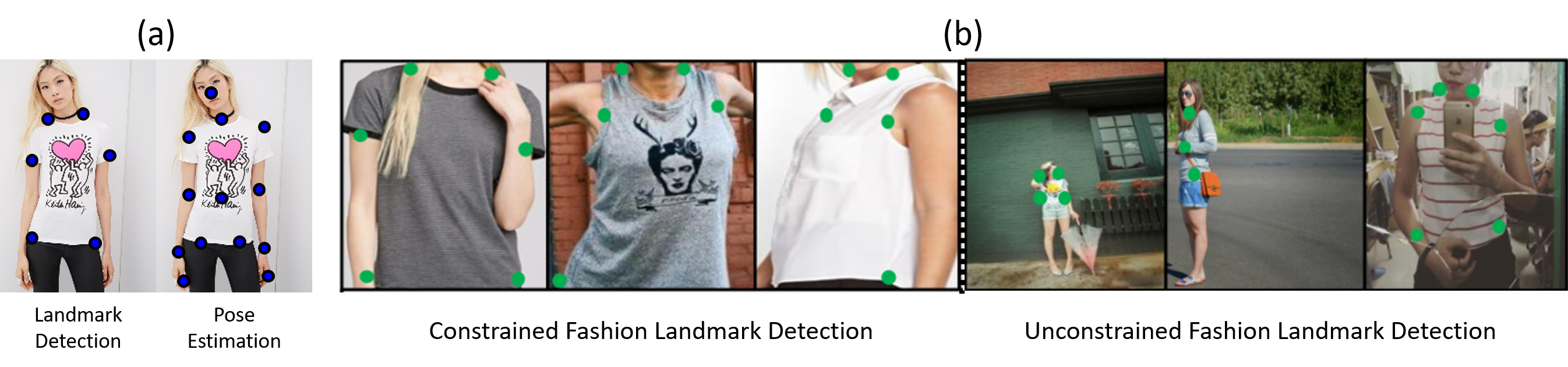}
  \caption{(a) The visual difference between landmark detection and pose estimation. (b) The visual difference between the constrained fashion landmark detection and the unconstrained fashion landmark detection~\cite{yan2017unconstrained}.}
  \label{fig:detection_comparison}
\end{figure}

It is worth mentioning the difference between fashion landmark detection and human pose estimation, which aims at locating human body joints as Fig.~\ref{fig:detection_comparison}(a) shows. 
Fashion landmark detection 
is a more challenging task than human pose estimation as the clothes are intrinsically more complicated than human body joints. In particular, garments undergo non-rigid deformations or scale variations, while human body joints usually have more restricted deformations. Moreover, the local regions of fashion landmarks exhibit more significant spatial and appearance variances than those of human body joints, as shown in Fig.~\ref{fig:detection_comparison}(b).

\subsubsection{State-of-the-art methods}

The concept of fashion landmark was first proposed by Liu~\emph{et al.}~\cite{liu2016deepfashion} in 2016, under the assumption that clothing bounding boxes are given as prior information in both training and testing. For learning the clothing features via simultaneously predicting the clothing attributes and landmarks, Liu~\emph{et al.} introduced FashionNet~\cite{liu2016deepfashion}, a deep model. The predicted landmarks were used to pool or gate the learned feature maps, which led to robust and discriminative representations for clothes. In the same year, Liu~\emph{et al.} also proposed a deep fashion alignment (DFA) framework~\cite{liu2016fashion}, which consisted of a three-stage deep convolutional network (CNN), where each stage subsequently refined previous predictions. Yan \emph{et al.}~\cite{yan2017unconstrained} further relaxed the clothing bounding box constraint, which is computationally expensive and inapplicable in practice. The proposed Deep LAndmark Network (DLAN) combined selective dilated convolution and recurrent spatial transformer, where bounding boxes and landmarks were jointly estimated and trained iteratively in an end-to-end manner. Both \cite{liu2016fashion} and \cite{yan2017unconstrained} are based on the regression model.

A more recent work \cite{wang2018attentive} indicated that the regression model is highly non-linear and difficult to optimize. Instead of regressing landmark positions directly, they proposed to predict a confidence map of positional distributions (\textit{i.e.}, heatmap) for each landmark. Additionally, they took into account the fashion grammar to help reason the positions of landmarks. For instance, ``\emph{left collar} $\leftrightarrow$ \emph{left waistline} $\leftrightarrow$ \emph{left hemline}'' that connecting in a human-parts kinematic chain was used as a constraint on the connected clothing parts to model the grammar topology. The human anatomical constraints were inherited in a recurrent neural network. Further, Lee~\emph{et al.}~\cite{Lee2019AGE} considered contextual knowledge of clothes and proposed a global-local embedding module for achieving more accurate landmark prediction performance. Ge~\emph{et al.}~\cite{DeepFashion2} presented a versatile benchmark Deepfashion2 for four tasks, clothes detection, pose estimation, human segmentation, and clothing retrieval, which covered most significant fashion detection works. They built a strong model Match R-CNN based on Mask R-CNN~\cite{maskrcnn2017} for solving the four tasks.

\subsubsection{Benchmark datasets}

As summarized in Table~\ref{tab:landmark_detection_datasets}, there are four benchmark datasets for fashion landmark detection, and the most used is Fashion Landmark Dataset~\cite{liu2016fashion}. These datasets differ in two major aspects: (1) standardization process of the images, and (2) pose and scale variations. 


\begin{table*}[t!]
\scriptsize
    \centering
        \caption{Summary of the benchmark datasets for fashion landmark detection task.}
        \vspace{-3mm}
    \begin{tabular}{l|c|c|c|c|c}
    \Xhline{1.0pt}
      \makecell[c]{Dataset name} & \makecell{Publish \\ time} & \makecell{\# of \\ photos} & \makecell{\# of \\landmark\\ annotations} & Key features & Sources \\
    \hline
        \makecell[l]{\href{http://mmlab.ie.cuhk.edu.hk/projects/DeepFashion/AttributePrediction.html}{DeepFashion-C}
        \label{refnote:deepFashionC}
        \cite{liu2016deepfashion}} & 2016 & 289,222 & 8 & \makecell[l]{Annotated with clothing bounding box, pose variation\\ type, landmark visibility, clothing type, category,\\ and attributes} & \makecell[l]{Online shopping\\ sites, Google \\Images}\\  
    \hline
        \makecell[l]{\href{http://mmlab.ie.cuhk.edu.hk/projects/DeepFashion/LandmarkDetection.html}{Fashion Landmark}\\ \href{http://mmlab.ie.cuhk.edu.hk/projects/DeepFashion/LandmarkDetection.html}{Dataset  (FLD)}
        ~\cite{liu2016fashion}} & 2016 & 123,016 & 8 & \makecell[l]{Annotated with clothing type, pose variation type, \\landmark visibility, clothing bounding box, and \\human body joint} & \makecell[l]{DeepFashion \\\cite{liu2016deepfashion}}\\
    \hline
        \makecell[l]{Unconstrained Land-\\mark Database~\cite{yan2017unconstrained}} & 2017 & 30,000 & 6 & \makecell[l]{The unconstrained datasets are often with cluttered \\background and deviate from image center.\\The visual comparison between constrained and\\ unconstrained dataset is presented in Fig.~\ref{fig:detection_comparison}(b)} & \makecell[l]{Fashion blogs\\ forums,  Deep-\\Fashion~\cite{liu2016deepfashion}}\\
    \hline
        \makecell[l]{\href{https://github.com/switchablenorms/DeepFashion2}{DeepFashion2}
        ~\cite{DeepFashion2}} & 2019 & 491,000 & \makecell{over 3.5x\\ of ~\cite{liu2016deepfashion}} & \makecell[l]{A versatile benchmark of four tasks including clothes\\ detection, pose estimation, segmentation, and  retrieval} & \makecell[l]{ Deep-Fashion \\\cite{liu2016deepfashion}, Online \\shopping sites}\\
    \Xhline{1.0pt}
    \end{tabular}
    \label{tab:landmark_detection_datasets}
\end{table*}

\subsubsection{Performance evaluations}

\begin{table*}[t!]
\notsotiny
\center
\caption{Performance comparisons of fashion landmark detection methods in terms of normalized error (NE).}
\vspace{-3mm}
\begin{tabular}{c|c|c|c|c|c|c|c|c|c|c}
  \Xhline{1.0pt}
  Dataset & Method & L. Collar & R. Collar & L. Sleeve & R. Sleeve & L. Waistline & R. Waistline & L. Hem & R. Hem & Avg. \\
  \hline
  \multirow{4}{*}{\makecell[l]{DeepFashion\\-C~\cite{liu2016deepfashion}}}
  & \makecell[l]{DFA~\cite{liu2016fashion}} & 0.0628 & 0.0637 & 0.0658 & 0.0621 & 0.0726 & 0.0702 & 0.0658 & 0.0663 & 0.0660 \\
  \cline{2-11}
  & \makecell[l]{DLAN~\cite{yan2017unconstrained}} & 0.0570 & 0.0611 & 0.0672 & 0.0647 & 0.0703 & 0.0694 & 0.0624 & 0.0627 & 0.0643 \\
  \cline{2-11}
  & \makecell[l]{AttentiveNet~\cite{wang2018attentive}} & 0.0415 & 0.0404 & 0.0496 & 0.0449 & 0.0502 & 0.0523 & 0.0537 & 0.0551 & 0.0484 \\
  \cline{2-11}
  & \makecell[l]{Global-Local~\cite{Lee2019AGE}} & \textbf{0.0312} & \textbf{0.0324} & \textbf{0.0427} & \textbf{0.0434} & \textbf{0.0361} & \textbf{0.0373} & \textbf{0.0442} & \textbf{0.0475} & \textbf{0.0393} \\
  \hline
  \hline
  \multirow{4}{*}{\makecell[l]{FLD~\cite{liu2016fashion}}} 
  & \makecell[l]{DFA~\cite{liu2016fashion}} & 0.0480 & 0.0480 & 0.0910 & 0.0890 & -- & -- & 0.0710 & 0.0720 & 0.0680 \\
  \cline{2-11}
  & \makecell[l]{DLAN~\cite{yan2017unconstrained}} & 0.0531 & 0.0547 & 0.0705 & 0.0735 & 0.0752 & 0.0748 & 0.0693 & 0.0675 & 0.0672 \\
  \cline{2-11}
  & \makecell[l]{AttentiveNet~\cite{wang2018attentive}} & 0.0463 & 0.0471 & \textbf{0.0627} & \textbf{0.0614} & 0.0635 & 0.0692 & 0.0635 & \textbf{0.0527} & 0.0583 \\
  \cline{2-11}
  & \makecell[l]{Global-Local~\cite{Lee2019AGE}} & \textbf{0.0386} & \textbf{0.0391} & 0.0675 & 0.0672 & \textbf{0.0576} & \textbf{0.0605} & \textbf{0.0615} & 0.0621 & \textbf{0.0568} \\
  \Xhline{1.0pt}
\end{tabular}
\label{table:landmark_detection_result}
    \begin{tablenotes}
      \scriptsize
      \item `L. Collar' represents left collar, while `R. Collar' represents right collar.
      \item ``--'' represents detailed results are not available.
    \end{tablenotes}
\end{table*}

Fashion landmark detection algorithms output the landmark (\textit{i.e.}, functional key point) locations in the clothing images. 
The normalized error (NE), which is defined as the $\ell_2$ distance between detected and the ground truth landmarks in the normalized coordinate space, is the most popular evaluation metric used in fashion landmark detection benchmarks.  
Typically, smaller values of NE indicates better results. 

We list the performance comparisons of leading methods on the benchmark datasets in Table~\ref{table:landmark_detection_result}. 
Moreover, the performances of the same method are different across datasets, but the rank is generally consistent.

\subsection{Fashion Parsing}

\begin{figure}[t]
  \centering
  \includegraphics[width=\linewidth]{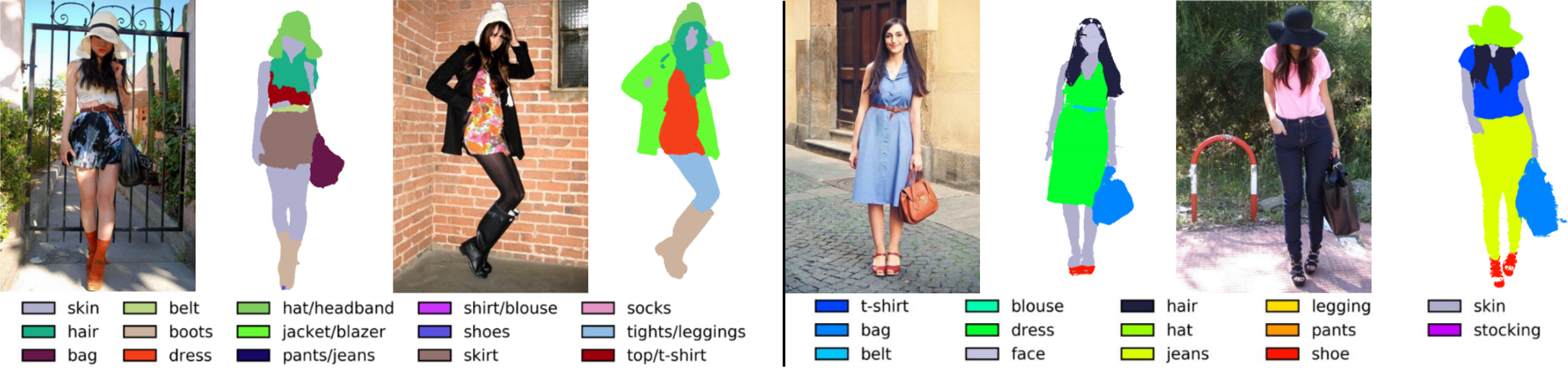}
  \caption{Examples of semantic segmentation for fashion images~\cite{ji:ijcai2018}.}
  \label{fig:fashion_parsing_example}
\end{figure}

Fashion parsing, human parsing with clothes classes, is a specific form of semantic segmentation, where the labels are based on the clothing items, such as dress or pants. Example of fashion parsing is shown in Fig.~\ref{fig:fashion_parsing_example}. Fashion parsing task distinguishes itself from general object or scene segmentation problems in that fine-grained clothing categorization requires higher-level judgment based on the semantics of clothing, the deforming structure within an image, and the potentially large number of classes.

\subsubsection{State-of-the-art methods}

The early work in fashion parsing was conducted by  
Yamaguchi~\emph{et al.}~\cite{KYamaguchi2012CVPR}. They exploited the relationship between clothing parsing and human pose estimation by refining two problems mutually. 
Specifically, clothing labels for every image segment were predicted with respect to body parts in a Conditional Random Field 
model. Then the predictions of clothing were incorporated as additional features for pose estimation. Their work, however, mainly focused on constrained parsing problem, where test images were parsed given user-provided tags indicating depicted clothing items. To overcome this limitation,~\cite{KYamaguchi2013, yamaguchi:tpami2014} proposed clothes parsing with a retrieval-based approach. For a given image, similar images from a parsed dataset were first retrieved, and then the nearest-neighbor parsings were transferred to the final result via dense matching. 
Since pixel-level labels required for model training were time-consuming, Liu \emph{et al.}~\cite{SLiu2014TMM} introduced  
the fashion parsing task with weak supervision from the color-category tags instead of pixel-level tags. They combined the human pose estimation module 
and (super)pixel-level category classifier learning module to generate category classifiers. They then applied the category tags to complete the parsing task.


Different from the abovementioned works 
that tended to consider the human pose first, which might lead to sub-optimal results due to the inconsistent targets between pose estimation and clothing parsing, recent research studies mainly attempted to relax this constraint. Dong~\emph{et al.}~\cite{dong2013deformable} proposed to use Parselets, a group of semantic image segments obtained from a low-level over-segmentation algorithm, as the essential elements. A Deformable Mixture Parsing Model (DMPM) based on the ``And-Or'' structure of sub-trees was built to jointly learn and infer the best configuration for both appearance and structure.  Next,~\cite{yang:cvpr2014, XLiang2016TMM} exploited contexts of clothing configuration, \textit{e.g.}, spatial locations and mutual relations of clothes items, to jointly parse a batch of clothing images given the image-level clothing tags. The proposed Clothes Co-Parsing (CCP) framework consists of two phases of inference: (1) image co-segmentation for extracting distinguishable clothes regions by applying exemplar-SVM classifiers, and (2) region co-labeling for recognizing garment items by optimizing a multi-image graphical model. Hidayati~\emph{et al.}~\cite{hidayati:mipr2019} integrated local features and possible body positions from each superpixel as the instances of the price-collecting Steiner tree problem.

In particular, the 
clothing parsing approaches 
based on hand-crafted processing steps need to be designed carefully to capture the complex correlations between clothing appearance and structure fully. To tackle this challenge, 
some CNN-based approaches have been explored. 
Liang~\emph{et al.}~\cite{liang:tpami2015} developed a framework based on active template regression to locate the mask of each semantic label, rather than assigning a label to each pixel. Two separate convolutional neural networks were utilized to build the end-to-end relation between the input image and the parsing result
. Following~\cite{liang:tpami2015}, Liang~\emph{et al.} later built a Contextualized 
CNN (Co-CNN) architecture~\cite{XLiang2015ICCV} to simultaneously capture the cross-layer context, global image-level context, and local super-pixel contexts to improve the accuracy of parsing results.


To address the issues of parametric and non-parametric human parsing methods that relied on the hand-designed pipelines composed of multiple sequential components, such as in~\cite{KYamaguchi2012CVPR, KYamaguchi2013, dong2013deformable, SLiu2014TMM}, Liu~\emph{et al.} presented a quasi-parametric human parsing framework~\cite{liu:cvpr2015}. The model inherited the merits of both parametric models and non-parametric models by the proposed Matching Convolutional Neural Network (M-CNN), which estimated the matching semantic region between the input image and KNN image. 
The works~\cite{Gong2017CVPR, liang:tpami2019} proposed self-supervised structure-sensitive learning approaches to explicitly enforce the consistency between the parsing results and the human joint structures. In this way, there is no need for specifically labeling human joints in model training.

Unlike previous approaches that only focused on single-person parsing task
,~\cite{Zhao2018MHP2, gong:eccv2018, CE2P2019} presented different methods for solving multi-person human parsing. Zhao~\emph{et al.}~\cite{Zhao2018MHP2} presented a deep Nested Adversarial Network\footnote{\url{https://github.com/ZhaoJ9014/Multi-Human-Parsing}} which contained three Generative Adversarial Network (GAN)-like sub-nets for semantic saliency prediction, instance-agnostic parsing, and instance-aware clustering respectively. These three sub-nets jointly learned in an end-to-end training way. Gong~\emph{et al.}~\cite{gong:eccv2018} designed a detection-free Part Grouping Network (PGN) to deal with multi-person human parsing in an image in a single pass. The proposed PGN integrates two twinned subtasks that can be mutually refined under a unified network, \textit{i.e.}, semantic part segmentation, and instance-aware edge detection. Further, Ruan~\emph{et al.}~\cite{CE2P2019} proposed CE2P framework\footnote{\url{https://github.com/liutinglt/CE2P}} containing three key modules, high resolution embedding module, global context embedding module, and edge perceiving module, for single human parsing. This work won the 1$^{st}$ place within all three human parsing tracks in the 2$^{nd}$ Look Into Person (LIP) Challenge\footnote{\url{https://vuhcs.github.io/vuhcs-2018/index.html}}. For multi-person parsing, they designed a global to local prediction process based on CE2P cooperating with Mask R-CNN to form M-CE2P framework and achieved the multi-person parsing goal.

In 2019, hierarchical graph was considered for human parsing tasks~\cite{3wayparsing2019, grapho2019}. Wang~\emph{et al.}~\cite{3wayparsing2019} defined the human body as a hierarchy of multi-level semantic parts and employed three processes (direct, top-down, and bottom-up) to capture the human parsing information for better parsing performance. For tackling human parsing in various domain via a single model without retraining on various datasets, Gong~\emph{et al.}~\cite{grapho2019} comprised hierarchical graph transfer learning based on the conventional parsing network to constitute a general human parsing model, Graphonomy\footnote{\url{https://github.com/Gaoyiminggithub/Graphonomy}}, which consisted of two processes. It first learned and propagated compact high-level graph representation among the labels within one dataset, and then transferred semantic information across multiple datasets. 



\subsubsection{Benchmark Datasets}

There are multiple datasets for fashion parsing, most of which are collected from Chictopia\footnote{\url{http://chictopia.com}}, a social networking website for fashion bloggers.  
Table~\ref{tab:clothing_parsing_datasets} summarizes the benchmark datasets for fashion parsing in more detail. To date, the most comprehensive one is the LIP dataset~\cite{Gong2017CVPR, liang:tpami2019}, containing over 50,000 annotated images with 19 semantic part labels captured from a wider range of viewpoints, occlusions, and background complexity. 

\begin{table*}[t!]
\renewcommand{\arraystretch}{1.1}
\notsotiny
    \centering
        \caption{Summary of the benchmark datasets for fashion parsing task.}
        \vspace{-3mm}
    \begin{tabular}{l|c|c|c|c|l|c}
    \Xhline{1.0pt}
        \multicolumn{2}{c|}{Dataset name} & \makecell{Publish \\ time} & \makecell{\# of \\ photos} & \makecell{\# of\\classes} & \makecell[c]{Key features} & Sources \\
    \hline
        \multicolumn{2}{c|}{\makecell[l]{\href{http://vision.is.tohoku.ac.jp/~kyamagu/research/clothing_parsing/}{Fashionista dataset}
        ~\cite{KYamaguchi2012CVPR}}} & 2012 & 158,235 & 56 & \makecell[l]{Annotated with tags, comments, and links} & \makecell[l]{Chictopia.com}\\ 
    \hline
        \multicolumn{2}{c|}{\makecell[l]{Daily Photos (DP)~\cite{dong2013deformable}}} & 2013 & 2,500 & 18 & \makecell[l]{High resolution images; Parselet definition} & \makecell[l]{Chictopia.com}\\
    \hline
        \multicolumn{2}{c|}{\makecell[l]{\href{http://vision.is.tohoku.ac.jp/~kyamagu/research/paperdoll/}{Paper Doll dataset}
        ~\cite{KYamaguchi2013, yamaguchi:tpami2014}}} & 2013 & 339,797 & 56 & \makecell[l]{Annotated with metadata tags denoting characteris-\\tics, \emph{e.g.}, color, style, occasion, clothing type, brand} & \makecell[l]{Fashionista~\cite{KYamaguchi2012CVPR}, \\ Chictopia.com}\\
    \hline
        \multicolumn{2}{c|}{\makecell[l]{\href{https://github.com/bearpaw/clothing-co-parsing}{Clothing Co-Parsing (CCP)}\\ \href{https://github.com/bearpaw/clothing-co-parsing}{SYSU-Clothes}
        ~\cite{yang:cvpr2014, XLiang2016TMM}}} & 2014 & 2,098 & 57 & \makecell[l]{Annotated with superpixel-level or image-level tags} & \makecell[l]{Online shopping websites}\\
    \hline
        \multicolumn{2}{c|}{\makecell[l]{\href{https://sites.google.com/site/fashionparsing/home}{Colorful-Fashion Dataset}\\ \href{https://sites.google.com/site/fashionparsing/home}{(CFD)}
        ~\cite{SLiu2014TMM}}} & 2014 & 2,682 & 23 & \makecell[l]{Annotated with 13 colors} & \makecell[l]{Chictopia.com}\\ 
    \hline 
        \multirow{3}{*}{\makecell[l]{ATR~\cite{liang:tpami2015}}} & \makecell[l]{Benchmark} & \multirow{3}{*}{2015} & 5,867 & \multirow{3}{*}{18} & \makecell[l]{Standing people in frontal/near-frontal view with good\\ visibilities of all body parts} & \makecell[l]{Fashionista~\cite{KYamaguchi2012CVPR}, Daily\\ Photos~\cite{dong2013deformable}, CFD~\cite{SLiu2014TMM}}\\
    \cline{2-2}\cline{4-4}\cline{6-7} 
        & \makecell[l]{Human Parsing\\ in the Wild}  &  & 1,833 &  & \makecell[l]{Annotated with pixel-level labels} & \makecell[l]{N/A}\\
    \hline
        \multicolumn{2}{c|}{\makecell[l]{Chictopia10k~\cite{XLiang2015ICCV}}} & 2015 & 10,000 & 18 & \makecell[l]{ It contains real-world images with arbitrary postures,\\ views and backgrounds} & \makecell[l]{Chictopia.com}\\
    \hline
        \multicolumn{2}{c|}{\makecell[l]{\href{http://hcp.sysu.edu.cn/li}{LIP}
        ~\cite{Gong2017CVPR, liang:tpami2019}}} & 2017 & 50,462 & 20 & \makecell[l]{Annotated with pixel-wise and body joints} & \makecell[l]{Microsoft COCO~\cite{lin:eccv2014}}\\
    \hline
        \multicolumn{2}{c|}{\makecell[l]{\href{https://sukixia.github.io/paper.html}{PASCAL-Person-Part}
        ~\cite{Pascal2017}}} & 2017 & 3,533 & 14 & \makecell[l]{It contains multiple humans per image in uncon-\\strained poses and occlusions} & \makecell[l]{N/A}\\
    \hline 
        \multirow{2}{*}{\makecell[l]{\href{https://lv-mhp.github.io/dataset}{MHP}
        }} & \makecell[l]{v1.0~\cite{li2017MHP1}} & 2017 & 4,980 & 18 & \makecell[l]{There are 7 body parts and 11 clothes and accessory\\ categories} & \makecell[l]{N/A}\\
    \cline{2-7} 
        & \makecell[l]{v2.0~\cite{Zhao2018MHP2}}  & 2018 & 25,403 & 58 & \makecell[l]{There are 11 body parts and 47 clothes and accessory\\ categories} & \makecell[l]{N/A}\\  
    \hline
        \multicolumn{2}{c|}{\makecell[l]{\href{http://sysu-hcp.net/lip/overview.php}{Crowd Instance-level Human}\\\href{http://sysu-hcp.net/lip/overview.php}{Parsing (CIHP)}
        ~\cite{gong:eccv2018}}} & 2018 & 38,280 & 19 & \makecell[l]{Multiple-person images; pixel-wise annotations in\\ instance-level} & \makecell[l]{Google, Bing}\\
    \hline
        \multicolumn{2}{c|}{\makecell[l]{\href{https://github.com/eBay/modanet}{ModaNet}
        ~\cite{zheng:mm2018}}} & 2018 & 55,176 & 13 & \makecell[l]{Annotated with pixel-level labels, bounding boxes,\\ and polygons} & \makecell[l]{PaperDoll~\cite{KYamaguchi2013}}\\
    \hline
        \multicolumn{2}{c|}{\makecell[l]{\href{https://github.com/switchablenorms/DeepFashion2}{DeepFashion2}
        ~\cite{DeepFashion2}}} & 2019 & 491,000 & 13 & \makecell[l]{A versatile benchmark of four tasks including clothes\\ detection, pose estimation, segmentation, and  retrieval.} & \makecell[l]{ DeepFashion~\cite{liu2016deepfashion},\\ Online shopping websites}\\
    \hline
        \multicolumn{2}{c|}{\makecell[l]{\href{https://fashionpedia.github.io/home/}{Fashionpedia}
        ~\cite{fashionpedia2020eccv}}} & 2020 & 48,000 & 46 & \makecell[l]{There are 294 fine-grained attributes. It contains high\\ resolution with 1710 $\times$ 2151 to maintain more details} & \makecell[l]{Flickr, Free license\\ photo websites}\\
    \Xhline{1.0pt}
    \end{tabular}
    \label{tab:clothing_parsing_datasets}
    \begin{tablenotes}
      \scriptsize
      \item N/A: there is no reported information to cite
    \end{tablenotes}
\end{table*}

\subsubsection{Performance Evaluations}

There are multiple metrics for evaluating fashion parsing methods: 
(1) average Pixel Accuracy (aPA) as the proportion of correctly labeled pixels in the whole image, (2) mean Average Garment Recall (mAGR), (3) Intersection over Union (IoU) as the ratio of the overlapping area of the ground truth and predicted area to the total area, (4) mean accuracy, (5) average precision, (6) average recall, (7) average F-1 score over pixels, and (8) foreground accuracy as the number of true pixels on the body over the number of actual pixels on the body.

In particular, most of the parsing methods 
are evaluated on Fashionista dataset~\cite{KYamaguchi2012CVPR} in terms of accuracy, average precision, average recall, and average F-1 score over pixels. We 
report the performance comparisons in Table 1 of Supplementary Material.

\subsection{Item Retrieval}
\label{subsec:item_retrieval}

As fashion e-commerce has grown over the years, there has been a high demand for innovative solutions to help customers find preferred fashion items with ease.
Though many fashion online shopping sites support keyword-based searches, there are many visual traits of fashion items that are not easily translated into words. It thus attracts tremendous attention from many research communities to develop cross-scenario image-based fashion retrieval tasks for matching the real-world fashion items to the online shopping image. Given a fashion image query, the goal of image-based fashion item retrieval is to find similar or identical items from the gallery.

\subsubsection{State-of-the-art methods}

The notable early work on automatic image-based clothing retrieval was presented by Wang and Zhang~\cite{wang2011clothes}. To date, extensive research studies have been devoted to addressing the problem of cross-scenario clothing retrieval since there is a large domain discrepancy between daily human photo captured in general environment and clothing images taken in ideal conditions (\textit{i.e.}, embellished photos used in online clothing shops). 
Liu~\emph{et al.}~\cite{liu2012street} proposed to utilize an unsupervised transfer learning method based on part-based alignment and features derived from the sparse reconstruction.  
Kalantidis~\emph{et al.}~\cite{kalantidis:icmr2013} presented clothing retrieval from the perspective of human parsing. A prior probability map of the human body was obtained through pose estimation to guide clothing segmentation, and then the segments were classified through locality-sensitive hashing. The visually similar items were retrieved by summing up the overlap similarities. Notably,~\cite{wang2011clothes, liu2012street, kalantidis:icmr2013} are based on hand-crafted features.



With the advances of deep learning, there has been a trend of building deep neural network architectures to solve the clothing retrieval task. 
Huang~\emph{et al.}~\cite{huang2015cross} developed a Dual Attribute-aware Ranking Network (DARN) to represent in-depth features using attribute-guided learning. DARN simultaneously embedded semantic attributes and visual similarity constraints into the feature learning stage, while at the same time modeling the discrepancy between domains. 
Li~\emph{et al.}~\cite{li:icpr2016} presented a hierarchical super-pixel fusion algorithm for obtaining the intact query clothing item and used sparse coding for improving accuracy. More explicitly, an over-segmentation hierarchical fusion algorithm with human pose estimation was utilized to get query clothing items and to retrieve similar images from the product clothing dataset. 

\begin{figure}[t!]
  \centering
  \includegraphics[width=\linewidth]{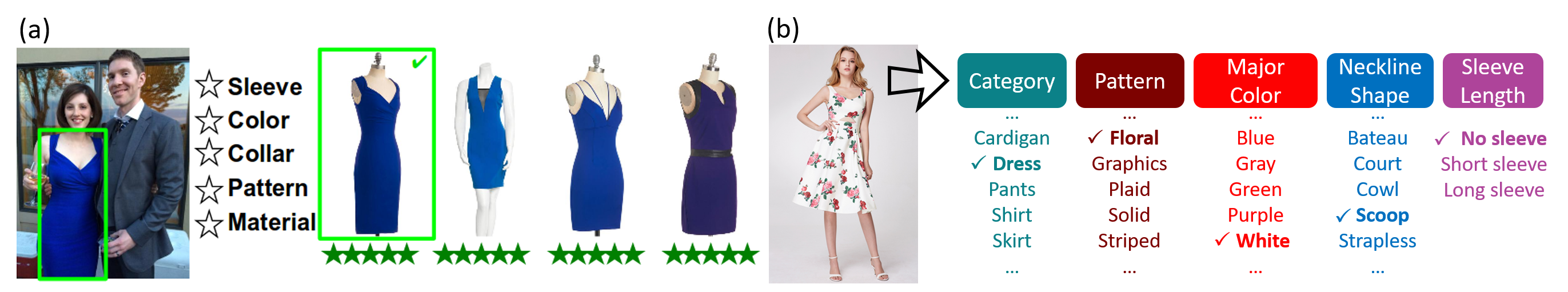}
  \caption{(a) An illustration of exact clothing retrieval~\cite{kiapour:iccv2015}. (b) An example for clothing attributes recognition.}
  \label{fig:_retrieval_attribute_example}
\end{figure}

The abovementioned studies 
are designed for similar fashion item retrieval. But more often, people desire to find the same fashion item, as illustrated in Fig.~\ref{fig:_retrieval_attribute_example}(a). The first attempt on this task was achieved 
by Kiapour~\emph{et al.}~\cite{kiapour:iccv2015}, who developed three different methods for retrieving the same fashion item in the real-world image from the online shop, the street-to-shop retrieval task. The three methods contained two deep learning baseline methods, and one method aimed to learn the similarity between two different domains, street and shop domains. 
For the same goal of learning the deep feature representation, Wang~\emph{et al.}~\cite{wang:icmr2016} adopted a Siamese network that contained two copies of the Inception-6 network with shared weights. Also, they introduced a \textit{robust contrastive loss} to alleviate over-fitting caused by some positive pairs (containing the same product) that were visually different, and used one multi-task fine-tuning approach to learn a better feature representation by tuning the parameters of the siamese network with product images and general images from ImageNet~\cite{imagenet2009}. Further, Jiang~\emph{et al.} ~\cite{jiang:mm2016} extended the one-way problem, street-to-shop retrieval task, to the bi-directional problem, street-to-shop and shop-to-street clothing retrieval task. They proposed a deep bi-directional cross-triplet embedding algorithm to model the similarity between cross-domain photos, and further expanded the utilization of this approach to retrieve a series of representative and complementary accessories 
to pair with the shop item~\cite{jiang:tomm2018}. 
Moreover, Cheng~\emph{et al.}~\cite{ZQCheng2017CVPR} increased the difficulty of image-based 
to video-based street-to-shop retrieval tasks
, which was more challenging because of the diverse viewpoint or motion blur
. They introduced three networks: 
Image Feature Network, Video Feature Network, and Similarity Network. They first did clothing detection and tracking for generating clothing trajectories. Then, Image Feature Network extracted deep visual features which would be fed into the 
LSTM framework for capturing the temporal dynamics in the Video Feature Network, and finally went to Similarity Network for pair-wise matching. 
For improving the existing algorithms for retrieval tasks, which only considered global feature vectors,~\cite{FashionRV2019} proposed a Graph Reasoning Network to build the similarity pyramid, which represented the similarity between a query and a gallery clothing by considering both global and local representation. 


\begin{figure}[t!]
  \centering
  \includegraphics[width=\linewidth]{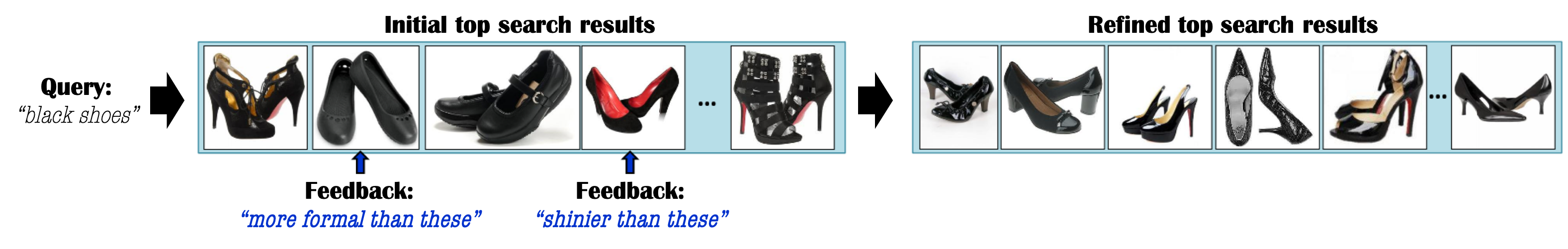}
  \caption{An example of interactive fashion item retrieval~\cite{kovashka:cvpr2012}.}
  \label{fig:clothing_retrieval_interactiveSearch}
\end{figure}

In the clothing retrieval methods described above, the retrieval is only based on query images that reflect users' needs, without considering that users may want to provide extra keywords to describe the desired attributes that are absent in the query image. Towards this goal, Kovashka~\emph{et al.} developed the WhittleSearch~\cite{kovashka:cvpr2012} 
that allows the user to upload a query image and provide additional relative attribute description as feedback for iteratively refining the image retrieval results (see Fig.~\ref{fig:clothing_retrieval_interactiveSearch}). Besides, for advancing the retrieval task with attribute manipulation, given a query image with attribute manipulation request, such as ``\textit{I want to buy a jacket 
like this query image, but with 
plush collar instead of round collar},'' the memory-augmented Attribute Manipulation Network (AMNet)~\cite{zhao:cvpr2017} updated the query representation encoding the unwanted attributes and replacing them to the desired ones. 
Later, Ak~\emph{et al.}
~\cite{ak:cvpr2018} presented the FashionSearchNet to learn regions and region-specific attribute representations by exploiting the attribute activation maps 
generated by the global average pooling 
layer. 
Different from~\cite{zhao:cvpr2017, ak:cvpr2018} that 
constructed visual representation of searched item by manipulating the visual representation of query image with the textual attributes in query text, \cite{laenen:wsdm2018} 
inferred the semantic relationship between visual and textual attributes in a joint multimodal embedding space. To help facilitate the reasoning of search results and user intent, Liao~\emph{et al.}~\cite{liao:mm2018} built EI (Exclusive and Independent) tree, that captured hierarchical structures of the fashion concepts
, which was generated by incorporating product hierarchies of online shopping websites and domain knowledge of fashion experts. 
It was then applied to guide the end-to-end deep learning procedure, mapping the deep implicit features to explicit fashion concepts.

As the fashion industry attracts much attention 
recently, there comes a multimedia grand challenge ``AI meets beauty''~\footnote{\url{https://challenge2020.perfectcorp.com/}}, which aimed for competing for the top fashion item recognition methods, held in ACM Multimedia yearly from 2018. Perfect Corp.
, CyberLink Corp., and National Chiao Tung University in Taiwan held this grand challenge and provided a large-scale image dataset 
of beauty and personal care products, namely the Perfect-500K dataset~\cite{perfect500k}. Lin~\emph{et al.}~\cite{MMGC2019} received the top performance in the challenge 
in 2019. They presented an unsupervised embedding learning to train a CNN model and combined the existing retrieval methods trained on different datasets to finetune the retrieval results.

\subsubsection{Benchmark datasets}

The existing clothing retrieval studies mainly focused on a cross-domain scenario. Therefore, most of the benchmark datasets were collected from daily photos and online clothing shopping websites. Table~\ref{tab:clothing_retrieval_datasets} gives a summary and link of the download page (if publicly available) of the benchmark datasets.

\begin{table*}[t]
\renewcommand{\arraystretch}{1.1}
\notsotiny
    \centering
        \caption{Summary of the benchmark datasets for fashion retrieval task.}
        \vspace{-3mm}
    \begin{tabular}{l|c|c|c|l|c}
    \Xhline{1.0pt}
        \multicolumn{2}{c|}{Dataset name} & \makecell{Publish \\ time} & \makecell{\# of \\ photos\\ (videos)}  & \makecell{Key features} & Sources \\
    \hline
        \multicolumn{2}{c|}{\makecell[l]{\href{http://skuld.cs.umass.edu/traces/mmsys/2014/user05.tar}{Fashion 10000}
        ~\cite{loni:mmsys2014}}} & 2014 &  32,398 & \makecell[l]{Annotated with 470 fashion categories
        } & \makecell[l]{Flickr.com}\\ 
    \hline
        \multicolumn{2}{c|}{\makecell[l]{Deep Search~\cite{huang:mm2014}}} & 2014 & 206,235 & \makecell[l]{With template matching, 7 attributes and type are extract-\\ed from the descriptions, i.e. pattern, sleeve, button panel,\\ collar, style, and color } & \makecell[l]{Taobao.com,\\ Tsmall.com,\\ Amazon.com}\\ 
    \hline
        \multicolumn{2}{c|}{\makecell[l]{DARN~\cite{huang2015cross}}} & 2015 & 545,373 & \makecell[l]{Online-offline upper-clothing image pair, annotated with\\ clothing attribute categories} & \makecell[l]{Online-shopping sites\\ and corresponding\\ customer review pages}\\ 
    \hline
        \multirow{2}{*}{\makecell[l]{\href{http://www.tamaraberg.com/street2shop}{Exact Street-}\\\href{http://www.tamaraberg.com/street2shop}{2Shop}
        ~\cite{kiapour:iccv2015}}} & \makecell[l]{Street photos} & \multirow{2}{*}{2015} & \makecell{20,357} & \multirow{2}{*}{\makecell[l]{39,479 pairs of exactly matching items worn in street\\ photos and shown in shop photos}} & \makecell[l]{ModCloth.com}\\ 
    \cline{2-2}\cline{4-4}\cline{6-6}
         & \makecell[l]{Shop photos} & & 404,683 & & \makecell[l]{Online clothing retailers}\\ 
    \hline 
        \multicolumn{2}{c|}{\makecell[l]{\href{http://mvc-datasets.github.io/MVC/}{MVC}
        ~\cite{liu:icmr2016}}} & 2016 & 161,638 & Annotated with 264 attribute labels & \makecell[l]{Online shopping sites}\\
    \hline
        \multirow{2}{*}{\makecell[l]{Li~\emph{et al.}~\cite{li:icpr2016}}} & \makecell[l]{Product Clothing} & \multirow{2}{*}{2016} & 15,690 & \multirow{2}{*}{\makecell[l]{Annotated with clothing categories}} & \makecell[l]{Online shopping sites}\\ 
        \cline{2-2}\cline{4-4}\cline{6-6}
        & \makecell[l]{Daily Clothing} &  & \makecell{4,206} & & \makecell[l]{Flickr.com}\\
    \hline
        \multicolumn{2}{c|}{\makecell[l]{\href{http://mmlab.ie.cuhk.edu.hk/projects/DeepFashion/InShopRetrieval.html}{DeepFashion (In-shop Clothes}\\ \href{http://mmlab.ie.cuhk.edu.hk/projects/DeepFashion/InShopRetrieval.html}{Retrieval Benchmark)}
        ~\cite{liu2016deepfashion}}} & 2016 & 52,712 & \makecell[l]{The resolution of images is 256$\times$256} & \makecell[l]{Online shopping sites\\ (Forever21 and Mogujie),\\ Google Images}\\
    \hline
        \multirow{2}{*}{\makecell[l]{Video2Shop\\\cite{ZQCheng2017CVPR}}} & \makecell[l]{Videos} & \multirow{2}{*}{2017} & \makecell{26,352\\(526)} & \multirow{2}{*}{\makecell[l]{39,479 exact matching pairs annotated with 14 categories\\ of clothes}} & \makecell[l]{Tmall MagicBox}\\ 
    \cline{2-2}\cline{4-4}\cline{6-6}
         & \makecell[l]{Online shopping} & & 85,677 & & \makecell[l]{Tmall.com, Taobao.com}\\ 
    \hline
        \multicolumn{2}{c|}{\makecell[l]{Dress like a star~\cite{garcia:iccvw2017}}} & 2017 & \makecell{7,000,000\\(40)} & \makecell[l]{It contains different movie genres such as animation, \\fantasy, adventure, comedy or drama} & \makecell[l]{YouTube.com}\\      
    \hline
        \multicolumn{2}{c|}{\makecell[l]{Amazon~\cite{liao:mm2018}}} & 2018 & 489,000 & \makecell[l]{Annotated with 200 clothing categories} & \makecell[l]{Amazon.com}\\ 
    \hline
        \multicolumn{2}{c|}{\makecell[l]{Perfect-500K~\cite{perfect500k}}} & 2018 & 500,000 & \makecell[l]{It is vast in scale, rich and diverse in content in order to\\ collect as many as possible beauty and personal care items} & \makecell[l]{e-commerce websites}\\
    \hline
        \multicolumn{2}{c|}{\makecell[l]{\href{https://github.com/switchablenorms/DeepFashion2}{DeepFashion2}
        ~\cite{DeepFashion2}}} & 2019 & 491,000 & \makecell[l]{a  versatile benchmark of four tasks including clothes\\ detection, pose estimation, segmentation, and  retrieval} & \makecell[l]{ DeepFashion~\cite{liu2016deepfashion},\\ Online shopping websites}\\
    \hline
        \multicolumn{2}{c|}{\makecell[l]{FindFashion~\cite{FashionRV2019}}} & 2019 & 565,041 & \makecell[l]{Merge two existing datasets, Street2Shop and Deep-\\Fashion, and label three attributes of the most affected} & \makecell[l]{Street2Shop~\cite{kiapour:iccv2015},\\DeepFashion~\cite{liu2016deepfashion}}\\
    \hline
        \multicolumn{2}{c|}{\makecell[l]{\href{https://github.com/Maryeon/asen}{Ma~\emph{et al.}}
        ~\cite{ma2020aaai}}} & 2020 & 180,000 & \makecell[l]{Since dataset for attribute-specific fashion retrieval is\\ lacking, this dataset rebuild three fashion dataset with\\ attribute annotations} & \makecell[l]{DARN~\cite{huang2015cross},\\FashionAI~\cite{Zou2019cvprw},\\DeepFashion~\cite{liu2016deepfashion}}\\
    \Xhline{1.0pt}
    \end{tabular}
    \label{tab:clothing_retrieval_datasets}
\end{table*}

\subsubsection{Performance evaluations}
\label{sec:item_retrieval_PE}
There are some evaluation metrics used to assess the performance of clothing retrieval methods. The different evaluation metrics used are as follows: (1) 
Top-\textit{k} retrieval accuracy, the ratio of queries with at least one matching item retrieved within the top-\textit{k} returned results, 
(2) 
Precision@\textit{k}, the ratio of items in the top-\textit{k} returned results that are matched with the queries, 
(3) 
Recall@\textit{k}, the ratio of matching items that are covered in the top-\textit{k} returned results, 
(4) 
Normalized Discounted Cumulative Gain (NDCG@\textit{k}), the relative orders among matching and non-matching items within the top-\textit{k} returned results, and 
(5) Mean Average Precision (MAP), which measures the precision of returned results at every position in the ranked sequence of returned results across all queries. 



Table 2 of the Supplementary Material presents the performance comparisons of some retrieval methods reviewed in this survey. We are unable to give comparisons for all different retrieval methods because the benchmarks they used are not consistent.

\section{Fashion Analysis}
\label{sec:fffashion_analysis}

Fashion is not only about what people are wearing but also reveals personality traits and other social cues. With immense potential in the fashion industry, precision marketing, sociological analysis, \emph{etc.}, intelligent fashion analysis on what style people choose to wear has thus gained increasing attention in recent years. In this section, we mainly focus on three fields of fashion analysis: attribute recognition, style learning, and popularity prediction. For each field, state-of-the-art methods, the benchmark datasets, and the performance comparison are summarised.

\subsection{Attribute Recognition}
\label{subsec:attribute_recognition}

Clothing attribute recognition is a multi-label classification problem that aims at determining which elements of clothing are associated with attributes among a set of $n$ attributes. As illustrated in Fig.~\ref{fig:_retrieval_attribute_example}(b), a set of attributes is a mid-level representation generated to describe the visual appearance of a clothing item. 



\subsubsection{State-of-the-art methods}

Chen \emph{et al.}~\cite{chen2012describing} learned a list of attributes for clothing on the human upper body. They extracted low-level features based on human pose estimation and then combined them for learning attribute classifiers. Mutual dependencies between the attributes capturing the rules of style (\textit{e.g.}, neckties are rarely worn with T-shirts) were explored by a Conditional Random Field (CRF) to make attribute predictions. A CRF-based model was also presented in~\cite{yamaguchi:iccv2015}. The model considered the location-specific appearance with respect to a human body and the compatibility of clothing items and attributes, which was trained using a max-margin learning framework.

Motivated by the large discrepancy between images captured in constrained and unconstrained environments, Chen \emph{et al.}~\cite{chen2015deep} studied a cross-domain attribute mining. They mined the data from \textit{clean} clothing images obtained from online shopping stores and then adapted it to unconstrained environments by using a deep domain adaptation approach. Lu~\emph{et al.}~\cite{lu:neurocomp2016} further presented a study on part-based clothing attribute recognition in a search and mining framework. The method consists of three main steps: (1) Similar visual search with the assistance of pose estimation and part-based feature alignment; (2) Part-based salient tag extraction that estimates the relationship between tags and images by the analysis of intra-cluster and inter-cluster of clothing essential parts; and (3) Tag refinement by mining visual neighbors of a query image. In the meantime, Li~\emph{et al.} \cite{li:eccv2016} learned to score the human body and attribute-specific parts jointly in a deep Convolutional Neural Network, and further improved the results by learning collaborative part modeling among humans and global scene re-scoring through deep hierarchical contexts. 

Different from the above methods that conducted attribute recognition only based on annotated attribute labels,~\cite{vittayakorn2016automatic, corbiere2017leveraging, han2017automatic} proposed to identify attribute vocabulary using weakly labeled image-text data from shopping sites. They used the neural activations in the deep network that generated attribute activation maps through training a joint visual-semantic embedding space to learn the characteristics of each attribute. In particular, Vittayakorn \emph{et al.}~\cite{vittayakorn2016automatic} exploited the relationship between attributes and the divergence of neural activations in the deep network. Corbiere \emph{et al.}~\cite{corbiere2017leveraging} trained two different and independent deep models to perform attribute recognition. Han~\emph{et al.}~\cite{han2017automatic} derived spatial-semantic representations for each attribute by augmenting semantic word vectors for attributes with their spatial representation. 

Besides, there are research works exploring attributes for recognizing the type of clothing items. Hidayati~\emph{et al.}~\cite{hidayati:mm2012} introduced clothing genre classification by exploiting the discriminative attributes of style elements, with an initial focus on the upperwear clothes. The work in~\cite{hidayati:tcyb2018} later extended \cite{hidayati:mm2012} to recognize the lowerwear clothes. Yu and Grauman~\cite{yu:cvpr2014} proposed a local learning approach for fine-grained visual comparison to predict which image is more related to the given attribute. Jia~\emph{et al.}~\cite{jia:aaai2016} introduced a notion of using the two-dimensional continuous image-scale space as an intermediate layer and formed a three-level framework, \textit{i.e.}, visual features of clothing images, image-scale space based on the aesthetic theory, and aesthetic words space consisting of words like ``formal'' and ``casual''. A Stacked Denoising Autoencoder Guided by Correlative Labels was proposed to map the visual features to the image-scale space. Ferreira~\emph{et al.}~\cite{ferreira2019iccv} designed a visual semantic attention model with pose guided attention for multi-label fashion classification.
Besides the clothing attribute recognition works, micro expression recognition is also an interesting task~\cite{ME2020mipr, ME2020acmmm}.

\subsubsection{Benchmark datasets}

There have been several clothing attribute datasets collected, as the older datasets are not capable of meeting the needs of the research goals. In particular, more recent datasets have a more practical focus. We summarize the clothing attribute benchmark datasets in Table~\ref{tab:attribute_recognition_datasets} and provide the links to download if they are available. 

\begin{table*}[t!]
\notsotiny
    \centering
        \caption{Summary of the benchmark datasets for clothing attribute recognition task.}
        \vspace{-3mm}
    \begin{tabular}{l|c|c|c|c|c|l|c}
    \Xhline{1.0pt}
        \multicolumn{2}{c|}{Dataset name} & \makecell{Publish \\ time} & \makecell{\# of \\ photos} & \makecell{\# of\\categories} & \makecell{\# of\\ attributes} & \makecell[c]{Key features} & Sources \\
    \hline
        \multicolumn{2}{c|}{\makecell[l]{\href{https://purl.stanford.edu/tb980qz1002}{Clothing Attributes}
        ~\cite{chen2012describing}}} & 2012 & 1,856 & 7 & 26 & \makecell[l]{Annotated with 23 binary-class attri-\\butes and 3 multi-class attributes} & \makecell[l]{Thesartorialist.com,\\Flickr.com}\\ 
    \hline
        \multicolumn{2}{c|}{\makecell[l]{Hidayati~\emph{et al.}~\cite{hidayati:mm2012}}} & 2012 & 1,077 & 8 & 5 & \makecell[l]{Annotated with clothing categories} & \makecell[l]{Online shopping sites}\\ 
    \hline
        \multicolumn{2}{c|}{\makecell[l]{\href{http://vision.cs.utexas.edu/projects/finegrained/}{UT-Zap50K shoe}
        ~\cite{yu:cvpr2014}}} & 2014 & 50,025  & N/A & 4 & \makecell[l]{Shoe images annotated with associated\\ metadata (shoe type, materials, gender,\\ manufacturer, \emph{etc}.)} & \makecell[l]{Zappos.com}\\ 
    \hline
        \multirow{4}{*}{\makecell[l]{Chen~\emph{et al.}\\\cite{chen2015deep}}} & \makecell[l]{Online-data} & \multirow{4}{*}{2015} & 341,021 & 15 & 67 & \makecell[l]{Each attribute has 1000+ images} &
        \makecell[l]{Online shopping sites}\\ 
    \cline{2-2}\cline{4-8}
        & \makecell[l]{Street-data-a} &  & 685 & N/A & N/A & \multirow{3}{*}{\makecell[l]{Annotated with fine-grained attributes}} & \makecell[l]{Fashionista~\cite{KYamaguchi2012CVPR}}\\
    \cline{2-2}\cline{4-6}\cline{8-8}
        & \makecell[l]{Street-data-b} &  & 8,000 & N/A & N/A &  & \makecell[l]{Parsing~\cite{dong:cvpr2014}}\\ 
    \cline{2-2}\cline{4-6}\cline{8-8}
        & \makecell[l]{Street-data-c} &  & 4,200 & N/A & N/A &  & \makecell[l]{Surveillance videos}\\ 
    \hline
        \multicolumn{2}{c|}{\makecell[l]{Lu~\emph{et al.}~\cite{lu:neurocomp2016}}} & 2016 & $\sim$1,1 M & N/A & N/A & \makecell[l]{Annotated with the associated tags} & \makecell[l]{Taobao.com}\\ 
    \hline
        \multicolumn{2}{c|}{\makecell[l]{\href{http://mmlab.ie.cuhk.edu.hk/projects/WIDERAttribute}{WIDER Attribute}
        ~\cite{li:eccv2016}}} & 2016 & 13,789 & N/A & N/A & \makecell[l]{Annotated with 14 human attribute la-\\bels and 30 event class labels} & \makecell[l]{The 50574 WIDER\\ images~\cite{xiong:cvpr2015}}\\ 
    \hline
        \multirow{2}{*}{\makecell[l]{Vittayakorn\\\emph{et al.}~\cite{vittayakorn2016automatic}}} & \makecell[l]{Etsy} & \multirow{2}{*}{2016} &  173,175 & N/A & 250 & \makecell[l]{Annotated with title and description of\\ the product} & \makecell[l]{Etsy.com}\\  
    \cline{2-2}\cline{4-8}
        & \makecell[l]{Wear} &  &  212,129 & N/A & \makecell{250} & \makecell[l]{Annotated with the associated tags} & \makecell[l]{Wear.jp}\\
    \hline
        \multicolumn{2}{c|}{\makecell[l]{\href{http://mmlab.ie.cuhk.edu.hk/projects/DeepFashion/AttributePrediction.html}{DeepFashion-C}
        ~\cite{liu2016deepfashion}}} & 2016 & 289,222 & 50 & 1,000 & \makecell[l]{Annotated with clothing bounding box,\\ type, category, and attributes} & \makecell[l]{Online shopping sites,\\ Google Images}\\  
    \hline
        \multicolumn{2}{c|}{\makecell[l]{\href{https://github.com/xthan/fashion-200k}{Fashion200K}
        ~\cite{han2017automatic}}} &  2017 & 209,544 & 5 & 4,404 & \makecell[l]{Annotated with product descriptions} & \makecell[l]{Lyst.com}\\ 
     \hline
        \multicolumn{2}{c|}{\makecell[l]{Hidayati~\emph{et al.}~\cite{hidayati:tcyb2018}}} & 2018 & 3,250 & 16 & 12 & \makecell[l]{Annotated with clothing categories} & \makecell[l]{Online shopping sites}\\ 
    \hline
        \multicolumn{2}{c|}{\makecell[l]{CatalogFashion-10x~\cite{Heilbron2019ICCV}}} & 2019 & 1M & 43 & N/A & \makecell[l]{The categories are identical to the\\ DeepFashion dataset~\cite{liu2016deepfashion}} & \makecell[l]{Amazon.com}\\ 
    \Xhline{1.0pt}
    \end{tabular}
    \label{tab:attribute_recognition_datasets}
    \begin{tablenotes}
      \scriptsize
      \item N/A: there is no reported information to cite
    \end{tablenotes}
\end{table*}

\subsubsection{Performance evaluations}

\begin{table*}[t!]
\notsotiny
\center
\caption{Performance comparisons of attribute recognition methods in terms of top-\textit{k} classification accuracy.}
\vspace{-3mm}
\begin{tabular}{c|c|c|c|c|c|c|c|c|c|c|c|c|c|c}
  \Xhline{1.0pt}
  \multirow{2}{*}{Method} & \multicolumn{2}{c|}{Category} & \multicolumn{2}{c|}{Texture} & \multicolumn{2}{c|}{Fabric} & \multicolumn{2}{c|}{Shape} & \multicolumn{2}{c|}{Part} & \multicolumn{2}{c|}{Style} & \multicolumn{2}{c}{All} \\
   & top-3 & top-5 & top-3 & top-5 & top-3 & top-5 & top-3 & top-5 & top-3 & top-5 & top-3 & top-5 & top-3 & top-5 \\
  \hline
  \makecell[l]{Chen \emph{et al.}~\cite{chen2012describing}} & 43.73 & 66.26 & 24.21 & 32.65 & 25.38 & 36.06 & 23.39 & 31.26 & 26.31 & 33.24 & 49.85 & 58.68 & 27.46 & 35.37 \\
  \hline
  \makecell[l]{DARN~\cite{huang2015cross}} & 59.48 & 79.58 & 36.15 & 48.15 & 36.64 & 48.52 & 35.89 & 46.93 & 39.17 & 50.14 & 66.11 & 71.36 & 42.35 & 51.95 \\
  \hline
  \makecell[l]{FashionNet~\cite{liu2016deepfashion}} &  82.58 & 90.17 & 37.46 & 49.52 & 39.30 & \textbf{49.84} & 39.47 & 48.59 & \textbf{44.13} & 54.02 & 66.43 & 73.16 & 45.52 & 54.61 \\
  \hline
  \makecell[l]{Corbiere \emph{et al.}~\cite{corbiere2017leveraging}} & 86.30 & 92.80 & \textbf{53.60} & 63.20 & 39.10 & 48.80 & 50.10 & 59.50 & 38.80 & 48.90 & 30.50 & 38.30 & 23.10 & 30.40 \\
  \hline
  \makecell[l]{AttentiveNet~\cite{wang2018attentive}} & \textbf{90.99} & \textbf{95.78} & 50.31 & \textbf{65.48} & \textbf{40.31} & 48.23 & \textbf{53.32} & \textbf{61.05} & 40.65 & \textbf{56.32} & \textbf{68.70} & \textbf{74.25} & \textbf{51.53} & \textbf{60.95} \\
  \Xhline{1.0pt}
\end{tabular}
    \begin{tablenotes}
      \scriptsize
      \item The best result is marked in bold. 
\end{tablenotes}
\label{table:attribute_recognition_result}
\end{table*}

Metrics that are used to evaluate the clothing attribute recognition models include the top-\textit{k} accuracy, mean average precision (MAP), and Geometric Mean (G-Mean). The top-\textit{k} accuracy and MAP have been described in Sec.~\ref{sec:item_retrieval_PE}, 
while G-mean measures the balance between classification performances on both the majority and minority classes. Most authors opt for a measure based on top-\textit{k} accuracy. We present the evaluation for general attribute recognition on the DeepFashion-C Dataset with different methods in Table~\ref{table:attribute_recognition_result}. 
The evaluation protocol is released in~\cite{liu2016deepfashion}. 

\subsection{Style Learning}

A variety of fashion styles is composed of different fashion designs. Inevitably, these design elements and their interrelationships serve as the powerful source-identifiers for understanding fashion styles. The key issue in this field is thus how to analyze discriminative features for different styles and also learn what style makes a trend.



\subsubsection{State-of-the-art methods}

An early attempt in fashion style recognition was presented by Kiapour~\emph{et al.}~\cite{MHKiapour2014}. They evaluated the concatenation of hand-crafted descriptors as style representation. Five different style categories are explored, including hipster, bohemian, pinup, preppy, and goth.

In the following studies~\cite{ESimo-Serra2016CVPR,SJiang2016AAAI, ma:aaai2017, vaccaro:uist2016, hsiao:iccv2017}, deep learning models were employed to represent fashion styles. In particular, Simo-Serra and Ishikawa~\cite{ESimo-Serra2016CVPR} developed a joint ranking and classification framework based on the Siamese network. The proposed framework was able to achieve outstanding performance with features being the size of a SIFT descriptor. To further enhance feature learning, Jiang \emph{et al.}~\cite{SJiang2016AAAI} used a consensus style centralizing auto-encoder (CSCAE) to centralize each feature of certain style progressively. Ma~\emph{et al.} introduced Bimodal Correlative Deep Autoencoder (BCDA)~\cite{ma:aaai2017}, a fashion-oriented multimodal deep learning based model adopted from Bimodal Deep Autoencoder~\cite{ngiam:icml2011}, to capture the correlation between visual features and fashion styles. The BCDA learned the fundamental rules of tops and bottoms as two modals of clothing collocations. The shared representation produced by BCDA was then used as input of the regression model to predict the coordinate values in the fashion semantic space that describes styles quantitatively. Vaccaro \emph{et al.} \cite{vaccaro:uist2016} presented a data-driven fashion model that learned the correspondences between high-level style descriptions (\textit{e.g.}, ``valentines day'' 
and low-level design elements (\textit{e.g.}, ``red cardigan'' 
by training polylingual topic modeling. This model adapted a natural language processing technique to learn latent fashion concepts jointly over the style and element vocabularies. Different from previous studies that sought coarse style classification, Hsiao and Grauman~\cite{hsiao:iccv2017} treated styles as discoverable latent factors by exploring style-coherent representation. An unsupervised approach based on polylingual topic models was proposed to learn the composition of clothing elements that are stylistically similar. Further, interesting work for learning the user-centric fashion information based on occasions, clothing categories, and attributes was introduced by Ma \emph{et al.}~\cite{ma2019acmmm}. Their main goal is to learn the information about ``what to wear for a specific occasion?'' from social media, \emph{e.g.}, Instagram. They developed a contextualized fashion concept learning model to capture the dependencies among occasions, clothing categories, and attributes.

\textbf{Fashion Trends Analysis}. A research pioneer in automatic fashion trend analysis was presented by Hidayati~\emph{et al.}~\cite{hidayati:mm2014}. They investigated fashion trends at ten different seasons of New York Fashion Week 
by analyzing the coherence (to occur frequently) and uniqueness (to be sufficiently different from other fashion shows) of visual style elements
. Following~\cite{hidayati:mm2014}, Chen~\emph{et al.}~\cite{chen:mm2015} utilized a learning-based clothing attributes approach to analyze the influence of the New York Fashion Show on people's daily life. Later, Gu~\emph{et al.}~\cite{gu:mm2017} presented QuadNet for analyzing fashion trends from street photos. The QuadNet was a classification and feature embedding learning network that consists of four identical CNNs, where the shared CNN was jointly optimized with a multi-task classification loss and a neighbor-constrained quadruplet loss. The multi-task classification loss aimed to learn the discriminative feature representation, while the neighbor-constrained quadruplet loss aimed to enhance the similarity constraint. A study on modeling the temporal dynamics of the popularity of fashion styles was presented in~\cite{he:www2016}. It captured fashion dynamics by formulating the visual appearance of items extracted from a deep convolutional neural network as a function of time.

Further, for statistics-driven fashion trend analysis,~\cite{chen:www2017} analyzed the best-selling clothing attributes corresponding to specific season via the online shopping websites statistics, e.g., (1) winter: gray, black, and sweater; (2) spring: white, red, and v-neckline. They designed a machine learning based method considering the fashion item sales information and the user transaction history for measuring the real impact of the item attributes to customers. Besides the fashion trend learning related to online shopping websites was considered, Ha~\emph{et al.}~\cite{ha:icwsm2017} used fashion posts from Instagram to analyze visual content of fashion images and correlated them with likes and comments from the audience. Moreover, Chang \emph{et al.} achieved an interesting work~\cite{YChang2017} on what people choose to wear in different cities. The proposed ``Fashion World Map'' framework exploited a collection of geo-tagged street fashion photos from an image-centered social media site, Lookbook.nu. They devised a metric based on deep neural networks to select the potential iconic outfits for each city and formulated the detection problem of iconic fashion items of a city as the prize-collecting Steiner tree (PCST) problem. A similar idea was proposed by Mall~\emph{et al.}~\cite{GeoStyle2019} in 2019; they established a framework\footnote{\url{https://github.com/kavitabala/geostyle}} for automatically analyzing the fashion trend worldwide in item attributes and style corresponding to the city and month. Specifically, they also analyzed how social events impact people wear, e.g., ``new year in Beijing in February 2014'' leads to red upper clothes.


Temporal estimation is an interesting task for general fashion trend analysis, the goal of which is to estimate when a style was made. Although visual analysis of fashion styles have been much investigated, this topic has not received much attention from the research community. 
Vittayakorn~\emph{et al.}~\cite{vittayakorn:wacv2017} proposed an approach to the temporal estimation task using 
CNN features and fine-tuning new networks to predict the time period of styles directly. This study also provided estimation analyses of what the temporal estimation networks have learned. 

\subsubsection{Benchmark datasets}

Table~\ref{tab:style_analysis_datasets} compares the benchmark datasets used in the literature. Hipster Wars
~\cite{MHKiapour2014} is the most popular dataset for style learning. We note that, instead of using the same dataset for training and testing, the study in~\cite{ESimo-Serra2016CVPR} used Fashion144k dataset~\cite{simoserra:cvpr2015} to train the model and evaluated it on Hipster Wars
~\cite{MHKiapour2014}. Moreover, the datasets related to fashion trend analysis 
were mainly collected from online media platforms during a specific time interval.

\begin{table*}[t!]
\notsotiny
    \centering
        \caption{Summary of the benchmark datasets for style learning task.}
        \vspace{-3mm}
    \begin{tabular}{l|c|c|c|l|c}
    \Xhline{1.0pt}
        \multicolumn{2}{c|}{Dataset name} & \makecell{Publish \\ time} & \makecell{\# of \\ photos} & \makecell[c]{Key features} & Sources \\
    \hline
        \multicolumn{2}{c|}{\makecell[l]{\href{http://tamaraberg.com/hipsterwars/}{Hipster Wars}~\cite{MHKiapour2014}
        }} & 2014 & 1,893 & \makecell[l]{Annotated with Bohemian, Goth, Hipster, Pinup,\\ or Preppy style} & \makecell[l]{Google Image Search}\\ 
    \hline
        \multicolumn{2}{c|}{\makecell[l]{Hidayati~\emph{et al.}~\cite{hidayati:mm2014}}} & 2014 & 3,276 & \makecell[l]{Annotated with 10 seasons of New York Fashion Week\\ (Spring/Summer 2010 to Autumn/Winter 2014)} & \makecell[l]{FashionTV.com}\\
    \hline
        \multicolumn{2}{c|}{\makecell[l]{\href{https://esslab.jp/~ess/en/data/fashion144k_stylenet/}{Fashion144k}~\cite{simoserra:cvpr2015}
        }} & 2015 & 144,169 & \makecell[l]{Each post contains text in the form of descriptions and gar-\\ment tags. It also consists of ``likes'' for scaling popularity} & \makecell[l]{Chictopia.com}\\
    \hline
        \multirow{3}{*}{\makecell[l]{Chen~\emph{et al.}~\cite{chen:mm2015}}} & \makecell[l]{Street-chic} & \multirow{2}{*}{2015} & 1,046 &  \makecell[l]{Street-chic images in New York from April 2014 to July\\ 2014 and from April 2015 to July 2015} & \makecell[l]{Flickr.com,\\ Pinterest.com}\\
        \cline{2-2}\cline{4-6}
        & \makecell[l]{New York\\ Fashion Show} &  & 7,914 & \makecell[l]{2014 and 2015 summer/spring New York Fashion Show} & \makecell[l]{Vogue.com}\\
    \hline
        \multicolumn{2}{c|}{\makecell[l]{Online Shopping ~\cite{SJiang2016AAAI}}} & 2016 & 30,000 & \makecell[l]{The 12 classes definitions are based on the fashion maga-\\zine~\cite{classesBased}} & \makecell[l]{Nordstrom.com,\\ barneys.com}\\
    \hline
        \multicolumn{2}{c|}{\makecell[l]{Fashion Data ~\cite{vaccaro:uist2016}}} & 2016 & 590,234 & \makecell[l]{Annotated with a title, text description, and a list of items\\ the outfit comprises} & \makecell[l]{Polyvore.com}\\
    \hline
        \multirow{2}{*}{\makecell[l]{He and McAuley\\\emph{et al.}~\cite{he:www2016}}} & \makecell[l]{Women} & \multirow{2}{*}{2016} & 331,173 &  \multirow{2}{*}{\makecell[l]{Annotated with users' review histories, time span between\\ March 2003 and July 2014}} & \makecell[l]{Amazon.com}\\
        \cline{2-2}\cline{4-4}
        & \makecell[l]{Men} &  & 100,654  &  & \\
    \hline
        \multirow{2}{*}{\makecell[l]{Vittayakorn~\emph{et al.}\\\cite{vittayakorn:wacv2017}}} & \makecell[l]{Flickr Clothing} & \multirow{2}{*}{2017} & 58,350 & \makecell[l]{Annotated with decade label and user provided metadata} & \makecell[l]{Flickr.com}\\
    \cline{2-2}\cline{4-6}
        & \makecell[l]{Museum Dataset} & & 9,421 & \makecell[l]{Annotated with decade label} & \makecell[l]{museum}\\
    \hline
        \multicolumn{2}{c|}{\makecell[l]{Street Fashion Style (SFS)~\cite{gu:mm2017}}} & 2017 & 293,105 & \makecell[l]{Annotated with user-provided tags, including geographical\\ and year information} & \makecell[l]{Chictopia.com}\\
    \hline
        \multicolumn{2}{c|}{\makecell[l]{Global Street Fashion\\(GSFashion)~\cite{YChang2017}}} & 2017 & 170,180 &  \makecell[l]{Annotated with (1) city name, (2) anonymized user ID,\\ (3) anonymized post ID, (4) posting date, (5) number of\\ likes of the post, and (6) user-provided metadata descri-\\bing the categories of fashion items along with the brand\\ names and the brand-defined product categories} & \makecell[l]{Lookbook.nu}\\
    \hline
        \multicolumn{2}{c|}{\makecell[l]{\href{https://pan.baidu.com/s/1boPm2OB}{Fashion Semantic Space (FSS)}
        ~\cite{ma:aaai2017}}} & 2017 & 32,133 & \makecell[l]{Full-body fashion show images; annotated with visual\\ features (\textit{e.g.}, collar shape, pants length, or color theme.)\\ and fashion styles (\textit{e.g.}, casual, chic, or elegant).} & \makecell[l]{Vogue.com}\\ 
    \hline
        \multicolumn{2}{c|}{\makecell[l]{\href{http://vision.cs.utexas.edu/projects/StyleEmbedding/}{Hsiao and Grauman}
        ~\cite{hsiao:iccv2017}}} &  2017 & 18,878 & \makecell[l]{Annotated with associated attributes} & \makecell[l]{Google Images}\\ 
    \hline
        \multicolumn{2}{c|}{\makecell[l]{Ha~\emph{et al.}~\cite{ha:icwsm2017}}} &  2017 & 24,752 & \makecell[l]{It comprises description of images, associated  metadata,\\ and annotated and predicted visual content variables} & \makecell[l]{Instagram}\\
    \hline
        \multicolumn{2}{c|}{\makecell[l]{Geostyle~\cite{GeoStyle2019}}} & 2019 & 7.7M &\makecell[l]{This dataset includes categories of 44 major world cities\\ across 6 continents, person body and face detection, and\\ canonical cropping} & \makecell[l]{Street Style~\cite{StreetStyle2017},\\Flickr100M~\cite{vittayakorn:wacv2015}}\\
    \hline
        \multicolumn{2}{c|}{\makecell[l]{FashionKE~\cite{ma2019acmmm}}} & 2019 & 80,629 &\makecell[l]{With the help of fashion experts, it contains 10 common\\ types of occasion concepts, e.g., dating, prom, or travel} & \makecell[l]{Instagram}\\
    \Xhline{1.0pt}
    \end{tabular}
    \label{tab:style_analysis_datasets}
    \begin{tablenotes}
      \scriptsize
      \item M means million.
    \end{tablenotes}
\end{table*}

\subsubsection{Performance evaluations}

The evaluation metrics used to measure the performance of existing fashion style learning models are precision, recall, and accuracy. 
Essentially, precision measures the ratio of retrieved results that are relevant, recall measures the ratio of relevant results that are retrieved, while accuracy measures the ratio of correct recognition. 
Although several approaches for fashion temporal analysis have been proposed, there is no extensive comparison between 
them. It is an open problem to determine which of the approaches perform better than others.

\subsection{Popularity Prediction.}
Precise fashion trend prediction is not only essential for fashion brands to strive for the global marketing campaign but also crucial for individuals to choose what to wear corresponding to the specific occasion. Based on the fashion style learning via the existing data (e.g., fashion blogs), it can better predict fashion popularity and further forecast future trend, which profoundly affect the fashion industry about multi-trillion US dollars.

\subsubsection{State-of-the-art methods}

Despite the active research in popularity prediction of online content on general photos with diverse categories~\cite{SMP2016aaai, SMP2017ijcai, hidayati:mm2017, massip:icmew2018, wen2018ijcai}, the popularity prediction on online social network specialized in fashion and style is currently understudied. The work in~\cite{yamaguchi:mm2014} presented a vision-based approach to quantitatively evaluate the influence of visual, textual, and social factors on the popularity of outfit pictures. They found that the combination of social and content information yields a good predictory for popularity. In~\cite{park:cscw2016}, Park~\emph{et al.} applied a widely-used machine learning algorithm to uncover the ingredients of success of fashion models and predict their popularity within the fashion industry by using data from social media activity. Recently, Lo~\emph{et al.}~\cite{lo2019icip} proposed a deep temporal sequence learning framework to predict the fine-grained fashion popularity of an outfit look.

Discovering the visual attractiveness has been the pursuit of artists and philosophers for centuries. Nowadays, the computational model for this task has been actively explored in the multimedia research community, especially with the focus on clothing and facial features. Nguyen~\emph{et al.}~\cite{nguyen:mm2012} studied how different modalities (\textit{i.e.}, face, dress, and voice) individually and collectively affect the attractiveness of a person. A tri-layer learning framework, namely Dual-supervised Feature-Attribute-Task 
network, was proposed to learn attribute models and attractiveness models simultaneously. Chen~\emph{et al.} \cite{chen:icme2013} focused on modeling fashionable dresses. The framework was based on two main components: basic visual pattern discovery using active clustering with humans in the loop, and latent structural SVM learning to differentiate fashionable and non-fashionable dresses. Next, Simo-Serra~\emph{et al.}~\cite{simoserra:cvpr2015} not only predicted the fashionability of a person's look on a photograph but also suggested what clothing or scenery the user should change to improve the look. For this purpose, they utilized a Conditional Random Field model to learn correlation among fashionability factors, such as the type of outfit and garments, the type of the user, and the scene type of the photograph.  

Meanwhile, the aesthetic quality assessment of online shopping photos is a relatively new area of study. A set of features related to photo aesthetic quality is introduced in~\cite{wang:icip2015}. To be more specific, in this work, Wang and Allebach investigated the relevance of each feature to aesthetic quality via the elastic net. They trained an SVM predictor with an optimal feature subset constructed by a wrapper feature selection strategy with the best-first search algorithm.

Other studies built computational attractiveness models to analyze facial beauty. A previous survey on this task was presented in \cite{liu:mtap2016}. Since some remarkable progress have been made on this subject, we extend \cite{liu:mtap2016} to cover recent advancements. 
Chen and Zhang~\cite{chen:neurocomp2016} introduced a causal effect criterion to evaluate facial attractiveness models. It proposed two-way measurements, \textit{i.e.}, by imposing interventions according to the model and by examining the change of attractiveness. To alleviate the need for rating history for the query, which prior works could not cope when there are none or few, Rothe~\emph{et al.}~\cite{rothe:cvpr2016} proposed to regress visual query to a latent space derived through matrix factorization for the known subjects and ratings. Besides, they employed a visual regularized collaborative filtering approach to infer inter-person preferences for attractiveness prediction. A psychologically inspired deep convolutional neural network (PI-CNN), which is a hierarchical model that facilitates both the facial beauty representation learning and predictor training, was later proposed in ~\cite{xu:iccasp2017}. To optimize the performance of the PI-CNN facial beauty predictor, \cite{xu:iccasp2017} introduced a cascaded fine-tuning scheme that exploits appearance features of facial detail, lighting, and color. To further consider the facial shape, Gao~\emph{et al.}~\cite{gao2018icpr} designed a multi-task learning framework that took appearance and facial shape into account simultaneously and jointly learned facial representation, landmark location, and facial attractiveness score. They proved that learning with landmark localization is effective for facial attractiveness prediction. For building flexible filters to learn the mapping adaptive for different attributes within a deep modal, Lin~\emph{et al.}~\cite{lin2019ijcai} proposed an attribute-aware convolutional neural network (AaNet) whose filter parameters were controlled adaptively by facial attributes. They also considered the cases without attribute labels and presented a pseudo attribute-aware network (P-AaNet), which learned to utilize image context information for generating attribute-like knowledge. Moreover, Shi~\emph{et al.}~\cite{shi2019icassp} introduced a co-attention learning mechanism that employed facial parsing masks for learning accurate representation of facial composition to improve facial attractiveness prediction.

There was a Social Media Prediction (SMP) challenge\footnote{\url{http://www.smp-challenge.com/}} held by Wu~\emph{et al.}~\cite{Wu2019SMPCA} in ACM Multimedia 2019 which aimed at the work focused on predicting future clicks of new social media posts before they were posted in social feeds. The participated teams need to build a new algorithm based on understanding and learning techniques and automatically predict popularity (formulated by clicks or visits) to achieve better performances. 

\textbf{Fashion Forecasting}.
There are strong practical interests in fashion sales forecasting, either by utilizing traditional statistical models~\cite{ni:expertsyst2011, choi:tcyb2012}, applying artificial intelligence models~\cite{banica:iffs2014}, or combining the advantages of statistics-based methods and artificial intelligence-based methods into hybrid models~\cite{kaya:iffs2014, ren:tcybsystems2015, chen:nca2017}. However, the problem of visual fashion forecasting, where the goal is to predict the future popularity of styles discovered from fashion images, has gained limited attention in the literature. In~\cite{alhalah:iccv2017}, Al-Halah~\emph{et al.} proposed to forecast the future visual style trends from fashion data in an unsupervised manner. The proposed approach consists of three main steps: (1) learning a representation of fashion images that captures clothing attributes using a supervised deep convolutional model; (2) discovering the set of fine-grained styles that are distributed across images using a non-negative matrix factorization framework; and (3) constructing styles' temporal trajectories  based on statistics of past consumer purchases for predicting the future trends.

\subsubsection{Benchmark datasets}
We summarize the benchmark datasets used to evaluate the popularity prediction models reviewed above in Table~\ref{tab:popularity_prediction_datasets}. There are datasets focused on face attractiveness called SCUT-FBP~\cite{xie:smc2015} and SCUT-FBP5500~\cite{liang2018SCUT}. Also, SMPD2019~\cite{Wu2019SMPCA} is specific for social media prediction.

\begin{table*}[t!]
\scriptsize
    \centering
        \caption{Summary of the benchmark datasets for popularity prediction task.}
        \vspace{-3mm}
    \begin{tabular}{c|c|c|c|c|c}
    \Xhline{1.0pt}
      \multicolumn{2}{c|}{Dataset name} & \makecell{Publish \\ time} & \makecell{\# of \\ photos} & Key features & Source \\
    \hline
        \multicolumn{2}{c|}{\makecell[l]{Yamaguchi~\emph{et al.}~\cite{yamaguchi:mm2014}}} & \makecell{2014} & 328,604 & \makecell[l]{Annotated with title, description, and user-provided labels} & \makecell[l]{Chictopia.com} \\
    \hline
        \multicolumn{2}{c|}{\makecell[l]{\href{http://www.hcii-lab.net/data/SCUT-FBP/EN/introduce.html}{SCUT-FBP}
        ~\cite{xie:smc2015}}} & \makecell{2015} & 500 & \makecell[l]{Asian female face images with attractiveness ratings} & \makecell[l]{Internet} \\
    \hline
        \multicolumn{2}{c|}{\makecell[l]{\href{https://esslab.jp/~ess/en/data/fashion144k_stylenet/}{Fashion144k}
        ~\cite{simoserra:cvpr2015}}} & 2015 & 144,169 & \makecell[l]{Each post contains text in the form of descriptions and garment\\ tags. It also consists of votes or ``likes'' for scaling popularity} & \makecell[l]{Chictopia.com}\\
    \hline
        \multirow{4}{*}{\makecell[l]{Park~\emph{et al.}\\\cite{park:cscw2016}}} & \makecell[l]{Fashion\\ Model\\ Directory} & \multirow{4}{*}{\makecell{2016}} & \multirow{4}{*}{\makecell{N/A}} & \makecell[l]{Profile of fashion models, including name, height, hip size, dress\\ size, waist size, shoes size, list of agencies, and details about all\\ runways the model walked on (year, season, and city)} & \makecell[l]{Fashionmodel-\\directory.com} \\
    \cline{2-2}\cline{5-6}
        & \makecell[l]{Instagram} &  &  & \makecell[l]{Annotated with the number of ``likes'' and comments, as well as the\\ metadata of the first 125 ``likes'' of each post} & \makecell[l]{Instagram} \\
    \hline
        \multicolumn{2}{c|}{\makecell[l]{TPIC17~\cite{SMP2017ijcai}}} & \makecell{2017} & 680,000 & \makecell[l]{With time information} & \makecell[l]{Flickr.com} \\
    \hline
        \multicolumn{2}{c|}{\makecell[l]{Massip~\emph{et al.}~\cite{massip:icmew2018}}} & \makecell{2018} & 6,000 & \makecell[l]{Using textual queries with hashtags of targeted image categories\\ to collect, e.g., \#selfie, or \#friend} & \makecell[l]{Instagram} \\
    \hline
        \multicolumn{2}{c|}{\makecell[l]{\href{https://github.com/HCIILAB/SCUT-FBP5500-Database-Release}{SCUT-FBP5500}
        ~\cite{liang2018SCUT}}} & \makecell{2018} & 5,500 & \makecell[l]{Frontal faces with various properties (gender, race, ages) and div-\\erse labels (face landmark and beauty score)} & \makecell[l]{Internet} \\
    \hline
        \multicolumn{2}{c|}{\makecell[l]{Lo~\emph{et al.}~\cite{lo2019icip}}} & \makecell{2019} & 380,000 & \makecell[l]{Within 52 different cities and the timeline was between 2008--2016} & \makecell[l]{lookbook.nu} \\
    \hline
        \multicolumn{2}{c|}{\makecell[l]{\href{http://smp-challenge.com/dataset}{SMPD2019}
        ~\cite{Wu2019SMPCA}}} & \makecell{2019} & 486,000 & \makecell[l]{It contains rich contextual information and annotations} & \makecell[l]{Flickr.com} \\
    \Xhline{1.0pt}
    \end{tabular}
    \label{tab:popularity_prediction_datasets}
    \begin{tablenotes}
      \scriptsize
      \item N/A: there is no reported information to cite.
    \end{tablenotes}
\end{table*}


\subsubsection{Performance evaluations}
Mean Absolute Percentage Error (MAPE), Mean Absolute Error (MAE), Mean Square Error (MSE), and Spearman Ranking Correlation (SRC) are the most used metrics to evaluate the popularity prediction performance. SRC is to measure the ranking correlation between groundtruth popularity set and predicted popularity set, varying from 0 to 1.

\section{Fashion Synthesis}
\label{sec:fffashion_synthesis}

Given an image of a person, we can imagine what that person would like in a different makeup or clothing style. We can do this by synthesizing a realistic-looking image. In this section, we review the progress to address this task, including style transfer, pose transformation, and physical simulation.

\subsection{Style Transfer}
Style transfer is transferring an input image into a corresponding output image such as transferring a real-world image into a cartoon-style image, transferring a non-makeup facial image into a makeup facial image, or transferring the clothing, which is tried on the human image, from one style to another. Style transfer in image processing contains a wide range of applications, \emph{e.g.}, facial makeup and virtual try-on. 

\subsubsection{State-of-the-art methods} The most popular style transfer work is pix2pix~\cite{isola2017image}, which is a general solution for style transfer. It learns not only the mapping from the input image to output image but also a loss function to train the mapping. For specific goal, based on a texture patch, ~\cite{jiang2017fashion, xian2018cvpr}\footnote{\url{https://github.com/janesjanes/Pytorch-TextureGAN}} transferred the input image or sketch to the corresponding texture. Based on the human body silhouette, ~\cite{lassner2017generative, Han2019iccv} inpainted compatible style garments to synthesize realistic images. An interesting work~\cite{Shi2019iccv} learned to transferred the facial photo into the game character style.


\textbf{Facial Makeup}. 
Finding the most suitable makeup for a particular human face is challenging, given the fact that a makeup style varies from face-to-face due to the different facial features. 
Studies on how to automatically synthesize the effects of with or without makeup on one's facial appearance have aroused interest recently. 
There is a survey on computer analysis of facial beauty provided 
in~\cite{laurentini:cviu2014}. However, it is limited to the context of the perception of attractiveness.

\begin{figure}[t!]
  \centering
  \includegraphics[width=\linewidth]{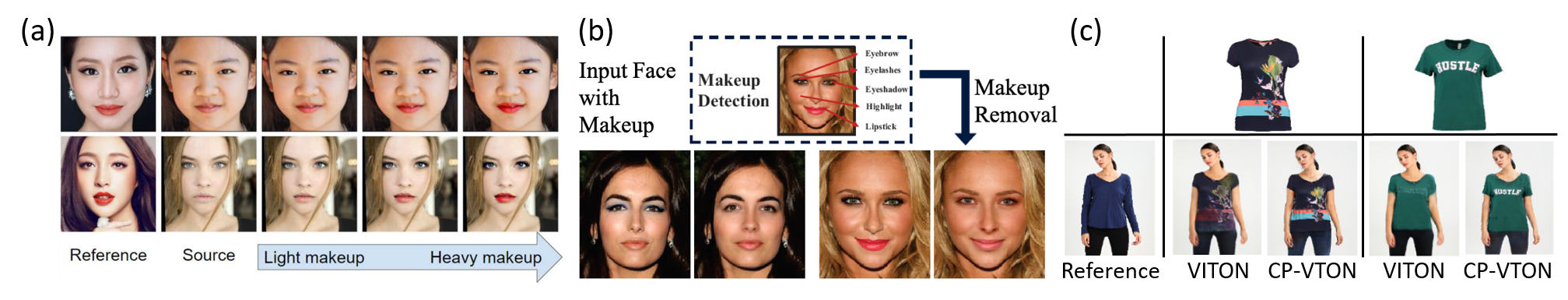}
  \vspace{-5mm}
  \caption{(a) Facial makeup transfer~\cite{chen:cvpr2019}. (b) Facial makeup detection and removal~\cite{AWang2016AAAI}. (c) Comparison of VITON~\cite{han2018viton} and CP-VTON~\cite{wang2018toward}.}
  \label{fig:style_transfer}
\end{figure}

Facial makeup transfer refers to translate the makeup from a given face to another one while preserving the identity as Fig.~\ref{fig:style_transfer}(a). It provides an efficient way for virtual makeup try-on and helps users select the most suitable makeup style. The early work achieved this task by image processing methods~\cite{CLi2015CVPR}, 
which decomposed images into multiple layers and transferred information from each layer after warping the reference face image to a corresponding layer of the non-makeup one. One major disadvantage of this method was that it required warping the reference face image to the non-makeup face image, which was very challenging and inclined to generate artifacts in many cases. Liu~\emph{et al.}~\cite{liu:ijcai2016} first adopted a deep learning framework for makeup transfer. They employed several independent networks to transfer each cosmetic on the corresponding facial part. The simple combination of several components applied in this framework, however, leads to unnatural artifacts in the output image. To address this issue, Alashkar~\emph{et al.}~\cite{alashkar:aaai2017} trained a deep neural network based makeup recommendation model from examples and knowledge base rules jointly. The suggested makeup style was then synthesized on the subject face. 

Different makeup styles result in significant facial appearance changes, which brings challenges to many practical applications, such as face recognition. Therefore, researchers also devoted themselves to makeup removal as Fig.~\ref{fig:style_transfer}(b), which is an ill-posed problem. Wang and Fu~\cite{AWang2016AAAI} proposed a makeup detector and remover framework based on locality-constrained dictionary learning. Li \emph{et al.}~\cite{YLi2018AAAI} later introduced a bi-level adversarial network 
architecture, where the first adversarial scheme was to reconstruct face images, and the second was to maintain face identity.


Unlike the aforementioned facial makeup synthesis methods that treat makeup transfer and removal as separate problems,~\cite{HChang2018CVPR, li2018beautygan, chen:cvpr2019, LADN2019iccv} performed makeup transfer and makeup removal simultaneously. Inspired by CycleGAN architecture~\cite{zhu2017unpaired}, Chang~\emph{et al.}~\cite{HChang2018CVPR} introduced the idea of asymmetric style transfer and a framework for training both the makeup transfer and removal networks together, each one strengthening the other. Li~\emph{et al.}~\cite{li2018beautygan} proposed a dual input/output generative adversarial network called BeautyGAN for instance-level facial makeup transfer. More recent work by Chen~\emph{et al.}~\cite{chen:cvpr2019} presented BeautyGlow that decomposed the latent vectors of face images derived from the Glow model into makeup and non-makeup latent vectors. For achieving better makeup and de-makeup performance, Gu~\emph{et al.}~\cite{LADN2019iccv} focused on local facial details transfer and designed a local adversarial disentangling network which contained multiple and overlapping local adversarial discriminators.

\textbf{Virtual Try-On}. Data-driven clothing image synthesis is a relatively new research topic that is gaining more and more attention. In the following, we review existing methods and datasets for addressing the problem of generating images of people in clothing by focusing on the styles. 
Han \emph{et al.}~\cite{han2018viton} utilized a coarse-to-fine strategy. Their framework, VIrtual Try-On Network (VITON), focused on trying an in-shop clothing image on a person image. It first generated a coarse tried-on result and predicted the mask for the clothing item. Based on the mask and coarse result, a refinement network for the clothing region was employed to synthesize a more detailed result. However,~\cite{han2018viton} fails to handle large deformation, especially with more texture details, due to the imperfect shape-context matching for aligning clothes and body shape. Therefore, a new model called Characteristic-Preserving Image-based Virtual Try-On Network (CP-VTON)~\cite{wang2018toward} was proposed. The spatial deformation can be better handled by a Geometric Matching Module, which explicitly aligned the input clothing with the body shape. The comparisons of VITON and CP-VTON are given in Fig.~\ref{fig:style_transfer}(c). There were several improved works~\cite{LAVITON2019iccv, Ayush2019iccv, VTNFP2019iccv, adaptive2020CVPR} based on CP-VTON. Different from the previous works which needed the in-shop clothing image for virtual try-on, FashionGAN~\cite{zhu2017your} and M2E-TON~\cite{M2E2019acmmm} presented target try-on clothing image based on text description and model image respectively. Given an input image and a sentence describing a different outfit, FashionGAN was able to ``redress'' the person. A segmentation map was first generated with a GAN according to the description. Then, the output image was rendered with another GAN guided by the segmentation map. M2E-TON was able to try on clothing from human$_A$ image to human$_B$ image, and two people can perform in different poses. Considering the runtime efficacy, Issenhuth \emph{et al.}~\cite{parsingfree2020eccv} proposed a parser free virtual try-on network. It designs a teacher-student architecture to free the parsing process during the inference time for improving efficiency.

Viewing the try-on performance from different views is also essential for virtual try-on task, Fit-Me~\cite{fitme2019icip} was the first work to do virtual try-on with arbitrary poses in 2019. They designed a coarse-to-fine architecture for both pose transformation and virtual try-on. Further, FashionOn~\cite{fashionon} applied the semantic segmentation for detailed part-level learning and focused on refining the facial part and clothing region to present more realistic results. They succeeded in preserving detailed facial and clothing information, perform dramatic posture, and also resolve the human limb occlusion problem in CP-VTON. Similar architecture to CP-VTON for virtual try-on with arbitrary poses was presented by~\cite{virtually2019acmmm}. They further made body shape mask prediction at the beginning of the first stage for pose transformation, and, in the second stage, they presented an attentive bidirectional GAN to synthesize the final result. For pose-guided virtual try-on, Dong~\emph{et al.}~\cite{Dong:iccv2019} further improved VITON and CP-VTON, which tackled the virtual try-on for different poses. Han~\emph{et al.}~\cite{clothflow2019iccv} proposed ClothFlow to focus on the clothing regions and model the appearance flow between source and target for transferring the appearance naturally and synthesizing novel result. Beyond the abovementioned image-based virtual try-on works, Dong~\emph{et al.}~\cite{FWGAN2019iccv} presented a video virtual try-on system, FWGAN, which learned to synthesize a video of virtual try-on results based on a person image, a target try-on clothing image and a series of target poses.
Increasing the image resolution of virtual try-on, \cite{unpaired2020cvpr,highRes2019iccvw} further design novel architectures for achieving multi-layer virtual try-on.

\subsubsection{Benchmark datasets}
Style transfer for fashion contains two hot tasks, facial makeup and virtual try-on. Existing makeup datasets applied for studying facial makeup synthesis typically assemble pairs of images for one subject: non-makeup image and makeup image pair. As for virtual try-on, since it is a highly diverse topic, there are several datasets for different tasks and settings. We summarize the datasets for style transfer in Table~\ref{tab:style_transfer_dataset}.

\begin{table*}[t!]
\notsotiny
\center
\caption{Summary of the benchmark datasets for style transfer task.}
\vspace{-3mm}
\begin{tabular}{l||c|c|c|c|c}
  \Xhline{1.0pt}
  \hline
  \makecell{Task} & Dataset name & \makecell{Publish\\time} & \makecell{\# of\\ photos} & Key features & Sources\\
  \hline
  \multirow{17}{*}{\makecell[l]{Facial\\ Makeup}} &
    \makecell[l]{Liu~\emph{et al.}~\cite{liu:ijcai2016}} & 2016 & 2,000 & \makecell[l]{1000 non-makeup faces and 1000 reference faces} & \makecell[l]{N/A} \\
  \cline{2-6}
    & \makecell[l]{Stepwise \\Makeup \cite{AWang2016AAAI}} & 2016 & 1,275 & \makecell[l]{Images in 3 sub-regions (
    eye
    , mouth
    , and 
    skin
    );\\ labeled with procedures of makeup} & \makecell[l]{N/A} \\
  \cline{2-6}
    & \makecell[l]{Beauty~\cite{zheng:fg2017}} & 2017 & 2,002 & \makecell[l]{1,001 subjects, where each subject has a pair of photos being with\\ and without makeup} & \makecell[l]{The Internet} \\
  \cline{2-6}
    & \makecell[l]{Before-After \\Makeup\cite{alashkar:aaai2017}} & 2017 & 1,922 & \makecell[l]{961 different females (224 Caucasian, 187 Asian, 300 African, and\\ 250 Hispanic) where one with clean face and another after pro-\\fessional makeup; annotated with facial attributes
    } & \makecell[l]{N/A}\\
  \cline{2-6}
    & \makecell[l]{Chang~\emph{et al.}\\\cite{HChang2018CVPR}} & 2018 & 2,192 & \makecell[l]{1,148 non-makeup images and 1,044 makeup images} & \makecell[l]{Youtube makeup\\ tutorial videos}\\
  \cline{2-6}
    & \makecell[l]{Makeup\\Transfer~\cite{li2018beautygan}} & 2018 & 3,834 & \makecell[l]{1,115 non-makeup images and 2,719 makeup images; assembled\\ with 5 different makeup styles (smoky-eyes, flashy, Retro, Ko-\\rean, and Japanese makeup styles), varying from subtle to heavy} & \makecell[l]{N/A}\\
  \cline{2-6}
    & \makecell[l]{\href{https://georgegu1997.github.io/LADN-project-page/}{LADN}
    ~\cite{LADN2019iccv}} & 2019 & 635 & \makecell[l]{333 non-makeup images and 302 makeup images} & \makecell[l]{The Internet}\\
 \cline{2-6}
    & \makecell[l]{\href{https://github.com/wtjiang98/PSGAN}{Makeup-Wild}~\cite{PSGAN2020cvpr}} & 2020 & 772 & \makecell[l]{369 non-makeup images and 403 makeup images} & \makecell[l]{The Internet}\\
  \hline\hline    
    \multirow{13}{*}{\makecell[l]{Virtual\\Try-On}} & \makecell[l]{\href{http://mmlab.ie.cuhk.edu.hk/projects/FashionGAN/}{LookBook}
    \\~\cite{yoo2016pixel}} & 2016 & 84,748 & \makecell[l]{The 9,732 top product images are associated with 75,016 fashion\\ model images} & \makecell[l]{Bongjashop, Jogunshop,\\ Stylenanda, SmallMan,\\ WonderPlace}\\
  \cline{2-6}
    & \makecell[l]{\href{http://mmlab.ie.cuhk.edu.hk/projects/DeepFashion/FashionSynthesis.html}{DeepFashion}
    \\\cite{liu2016deepfashion}} & 2016 & 78,979 & \makecell[l]{The corresponding upper-body images, sentence descriptions, and\\ human body segmentation maps} & \makecell[l]{Forever21} \\
  \cline{2-6}  
    & \makecell[l]{CAGAN~\cite{CAGAN2017iccvw}} & 2017 & 15,000 & \makecell[l]{Frontal view human images and paired upper-body garments\\ (pullovers and hoodies)} & \makecell[l]{Zalando.com} \\
  \cline{2-6}  
    & \makecell[l]{\href{https://github.com/xthan/VITON}{VITON}
    ~\cite{han2018viton}} & 2018 & 32,506 & \makecell[l]{Pairs of frontal-view women and top clothing images} & \makecell[l]{N/A} \\
  \cline{2-6}  
    & \makecell[l]{\href{https://fashiontryon.wixsite.com/fashiontryon}{FashionTryOn}
    \\\cite{virtually2019acmmm}} & 2019 & 28,714 & \makecell[l]{Pairs of same person with same clothing in 2 different poses and\\ one corresponding clothing image} & \makecell[l]{Zalando.com} \\
  \cline{2-6}  
    & \makecell[l]{FashionOn~\cite{fashionon}} & 2019 & 22,566 & \makecell[l]{Pairs of same person with same clothing in 2 different poses and\\ one corresponding clothing image} & \makecell[l]{The Internet,\\DeepFashion~\cite{liu2016deepfashion}} \\
  \cline{2-6}  
    & \makecell[l]{Video Virtual\\ Try-on \cite{FWGAN2019iccv}} & 2019 & \makecell{791\\videos} & \makecell[l]{Each video contains 250--300 frame numbers} & \makecell[l]{fashion model catwalk} \\
  \hline
  \Xhline{1.0pt}
\end{tabular}
\label{tab:style_transfer_dataset}
    \begin{tablenotes}
      \scriptsize
      \item N/A: there is no reported information to cite.
    \end{tablenotes}
\end{table*}

\subsubsection{Performance evaluations}
\label{sec:virtual_try_on_PE}
The evaluation for 
style transfer is generally based on subjective assessment or user study. That is, the participants rate the results into some certain degrees, such as ``Very bad'', ``Bad'', ``Fine'', ``Good'', and ``Very good''. The percentages of each degree are then calculated to quantify the quality of results. 
Besides, there are objective comparisons for virtual try-on, in terms of inception score (IS) or structural similarity (SSIM). IS is used to evaluate the synthesis quality of images quantitatively. The score will be higher if the model can produce visually diverse and semantically meaningful images. On the other hand, SSIM is utilized to measure the similarity between the reference image and the generated image ranging from zero (dissimilar) to one (similar).


\subsection{Pose Transformation}
Given a reference image and a target pose only with keypoints, the goal of pose transformation is to synthesize pose-guided person image in different posture while keeping personal information. A few examples of pose transformation are shown in Fig.~\ref{fig:pose_transformation_example}. Pose transformation, in particular, is a challenging task since the input and output are not spatially aligned.


\subsubsection{State-of-the-art methods}
\begin{figure}[t!]
  \centering
  \includegraphics[width=\linewidth]{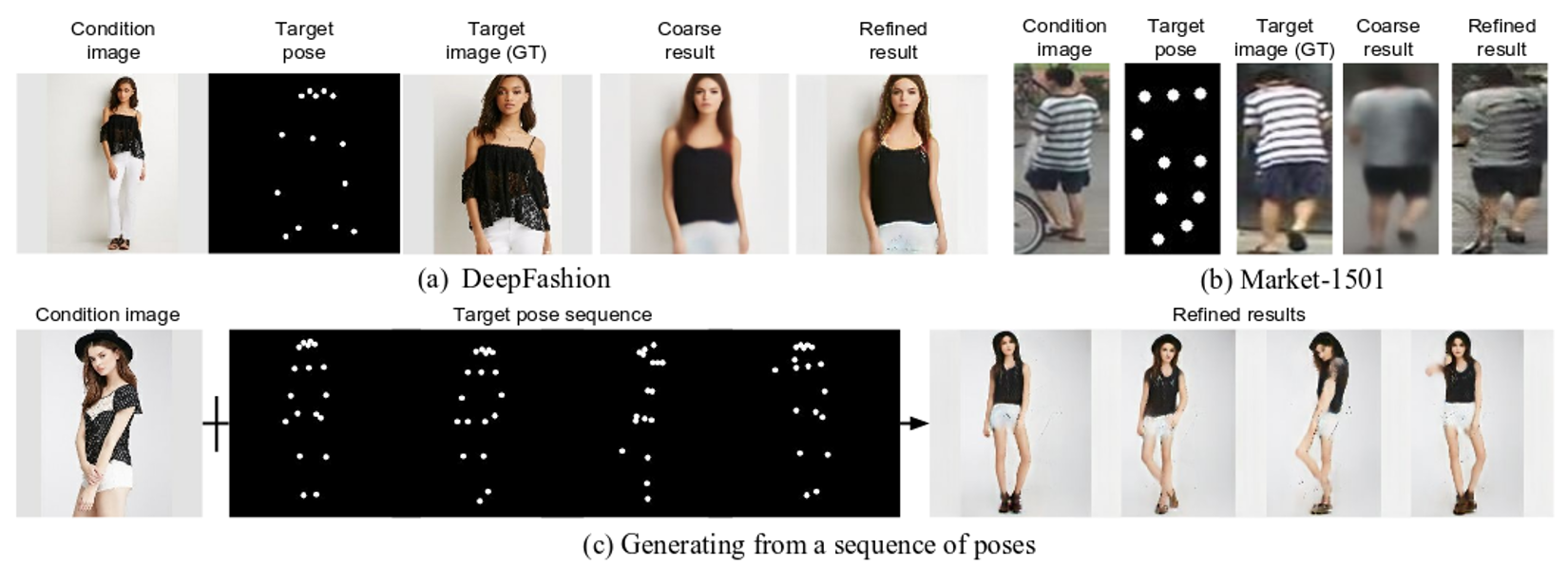}\\
  \caption{Examples of pose transformation~\cite{ma2017pose}.}
  \label{fig:pose_transformation_example}
\end{figure}

A two-stage adversarial network PG$^2$~\cite{ma2017pose} achieved an early attempt on this task. A coarse image under the target pose was generated in the first stage and then refined in the second stage. The intermediate results and final results are shown 
in Fig.~\ref{fig:pose_transformation_example}(a)-(b) 
with two benchmark datasets. 
However, the results were highly blurred, especially for texture details. To tackle the problem, the affine transform was employed to keep textures in the generated images better
. Siarohin~\emph{et al.}~\cite{siarohin2018deformable} designed a deformable GAN, in which the key deformable skip elegantly transforms high-level features for each body part. Similarly, the work in~\cite{balakrishnan2018synthesizing} employed body part segmentation masks to guide the image generation. The proposed framework contained four modules, including source image segmentation, spatial transformation, foreground synthesis, and background synthesis that can be trained jointly. 
Further, Si~\emph{et al.}~\cite{Si:cvpr2018} introduced a multistage pose-guided image synthesis framework, which divided the network into three stages for pose transform in a novel 2D view, foreground synthesis, and background synthesis. 

To break the data limitation of previous studies, Pumarola~\emph{et al.}~\cite{pumarola2018unsupervised} borrowed the idea from~\cite{zhu2017unpaired} by leveraging cycle consistency
. In the meantime, the works in~\cite{ma2018disentangled,esser2018variational} formulated the problem 
from the perspective of variational auto-encoder (VAE). 
They can successfully model the body shape; however, their results were less faithful to the appearance of reference images since they generated results from highly compressed features sampled from the data distribution. To improve the appearance performance
, Song~\emph{et al.}~\cite{song2019pose}\footnote{\url{https://github.com/SijieSong/person_generation_spt}} designed a novel pathway to decompose the hard mapping into two accessible subtasks, semantic parsing transformation and appearance generation. Firstly, for simplifying the non-rigid deformation learning, it transformed the posture in semantic parsing maps. Then, synthesizing the semantic-aware human information to the previous synthesized semantic maps formed the realistic final results. 
Conducting the concept of optical flow, Ren~\emph{et al.}~\cite{gfla2020cvpr} proposed a differentiable global-flow local-attention framework to ensemble the features between source human, source pose, and target pose.

\subsubsection{Benchmark datasets}
The benchmark datasets for pose transformation are very limited. The most used two benchmark datasets are Market-1501~\cite{zheng2015scalable} and DeepFashion (In-shop clothes retrieval benchmark)~\cite{liu2016deepfashion}. Besides, there is one dataset collected in videos~\cite{balakrishnan2018synthesizing}. All of the three datasets are summarized in Table~\ref{table:pose_transformation_dataset}.

\begin{table*}[t!]
\scriptsize
\center
\caption{Summary of the benchmark datasets for pose transformation task.}
\vspace{-3mm}     
\begin{tabular}{c|c|c|c|c}
  \Xhline{1.0pt}
   Dataset name & \makecell{Publish\\time} & \makecell{\# of\\ photos} & Key features & Sources\\
   \hline
    \makecell[l]{\href{http://vision.imar.ro/human3.6m}{Human3.6M}
    ~\cite{human36m_tpami}} & 2014 & 3.6M & \makecell[l]{ It contains 32 joints for each skeleton} & \makecell[l]{self-collected}\\
  \hline 
    \makecell[l]{\href{http://www.liangzheng.com.cn/Project/project_reid.html}{Market-1501}
    ~\cite{zheng2015scalable}} & 2015 & 32,668 & \makecell[l]{Multiple viewpoints; the resolution of\\ images is 128$\times$64} & \makecell[l]{A supermarket in Tsinghua University}\\
  \hline
    {\makecell[l]{\href{http://mmlab.ie.cuhk.edu.hk/projects/DeepFashion/InShopRetrieval.html}{DeepFashion}
    ~\cite{liu2016deepfashion}}} & 2016 & 52,712 & \makecell[l]{The resolution of images is 256$\times$256} & \makecell[l]{Online shopping sites (Forever21 and \\Mogujie), Google Images}\\
  \hline
    \makecell[l]{Balakrishnan~\emph{et al.}~\cite{balakrishnan2018synthesizing}} & 2018 & N/A & \makecell[l]{ Action classes: golf swings (136 videos),\\ yoga/workout routines (60 videos), and \\tennis actions (70 videos)} & \makecell[l]{YouTube}\\
  \Xhline{1.0pt}
\end{tabular}
    \begin{tablenotes}
      \scriptsize
      \item N/A: there is no reported information to cite. M means million.
    \end{tablenotes}
\label{table:pose_transformation_dataset}
\end{table*}

\subsubsection{Performance evaluations}
There are objective comparisons for pose transformation, in terms of 
IS and SSIM, which have been introduced in Sec.~\ref{sec:virtual_try_on_PE}. Additionally, to eliminate the effect of background, mask-IS and mask-SSIM were proposed in~\cite{ma2017pose}. 

\subsection{Physical Simulation}
\begin{figure}[t!]
  \centering
  \includegraphics[width=\linewidth]{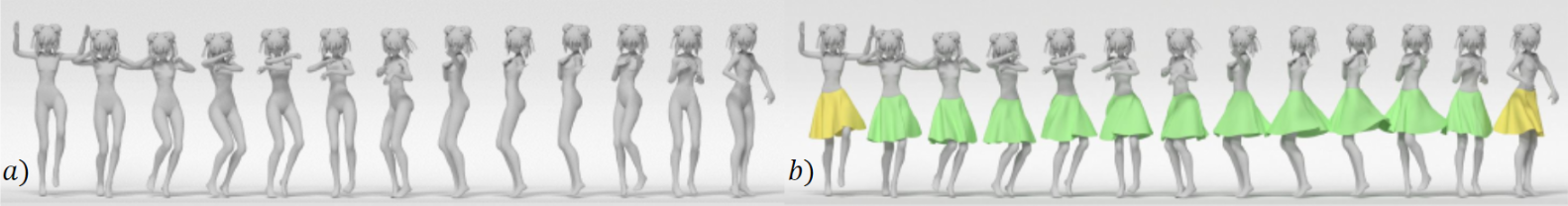}\\
  \caption{Examples of physical simulation~\cite{wang2019acmtog}.}
  \label{fig:physical_simulation_example}
\end{figure}

For more vivid fashion synthesis performance, physical simulation plays a crucial role. The abovementioned synthesis works are within the 2D domain, limited in the simulation of the physical deformation, e.g., shadow, pleat, or hair details. For advancing the synthesis performance with dynamic details (drape, or clothing-body interactions), there are physical simulation works based on 3D information. Take Fig.~\ref{fig:physical_simulation_example} for example. Based on a body animation sequence (a), the shape of the target garment and keyframes (marked by yellow), Wang~\emph{et al.}~\cite{wang2019acmtog} learned the intrinsic physical properties and simulated to other frames with different postures which are shown in (b).

\subsubsection{State-of-the-art methods}
The traditional pipeline for designing and simulating realistic clothes is to use computer graphics to build 3D models and render the output images~\cite{wang:tog11, guan:tog2012, yang:iccv2017, pons-moll:tog2017}. 
For example, Wang~\emph{et al.}~\cite{wang:tog11} developed a piecewise linear elastic model for learning both stretching and bending in real cloth samples. For learning the physical properties of clothing on different human body shapes and poses, Guan~\emph{et al.}~\cite{guan:tog2012} designed a pose-dependent model to simulate clothes deformation. For simulating regular clothing on fully dressed people in motion, Pons-Moll~\emph{et al.}~\cite{pons-moll:tog2017} designed a multi-part 3D model called ClothCap. Firstly, it separated different garments from the human body for estimating the clothed body shape and pose under the clothing. Then, it tracked the 3D deformations of the clothing over time from 4D scans to help simulate the physical clothing deformations in different human posture. To enhance the realism of the garment on human body, L{\"{a}}hner~\emph{et al.}~\cite{Zorah2018DeepWrinkles} proposed a novel framework, which composed of two complementary modules: (1) A statistical model learned to align clothing templates based on 3D scans of clothed people in motion and a linear subspace model factored out the human body shape and posture. (2) A cGAN added high-resolution geometric details to normal maps and simulated the physical clothing deformations. 

For advancing the physical simulation with non-linear deformations of clothing, Santesteban~\emph{et al.}~\cite{Santesteban2019cgf} presented a two-level learning-based clothing animation method for highly efficient virtual try-on simulation. There were two fundamental processes: it first applied global body-shape-dependent deformations to the garment and then predicted dynamic wrinkle deformations based on the body shape and posture. Further, Wang~\emph{et al.}~\cite{wang2019acmtog} introduced a semi-automatic method for authoring garment animation, which first encoded essential information of the garment shape and based on the intrinsic garment representation and target body motion, it learned to reconstruct garment shape with physical properties automatically. Based only on a single-view image, Yang~\emph{et al.}~\cite{yang:iccv2017} proposed a method to recover a 3D mesh of garment with the 2D physical deformations. Given a single-view image, a human-body database, and a garment-template database as input, it first preprocessed with garment parsing, human body reconstruction, and features estimation. Then, it synthesized the initial garment registration and garment parameter identification for reconstructing body and clothing models with physical properties. Besides, Yu~\emph{et al.}~\cite{yu2019cvpr} enhanced the simulation performance with a two-step model called SimulCap, which combined the benefits of capture and physical simulation. The first step aimed to get a multi-layer avatar via double-layer reconstruction and multi-layer avatar generation. Then, it captured the human performance by body tracking and cloth tracking for simulating the physical clothing-body interactions.


\subsubsection{Benchmark datasets}
Datasets for physical simulation are different from other fashion tasks since the physical simulation is more related to computer graphics than computer vision. Here, we discuss the types of input data used by most physical simulation works. Physical simulation working within the fashion domain focus on clothing-body interactions, and datasets can be categorized into real data and created data. For an example of real data, for each type of clothing,
~\cite{Zorah2018DeepWrinkles} captured 4D scan sequences at 60 fps 
in motion and dressed in a full-body suit. As for created data,
~\cite{Santesteban2019cgf} was based on one garment to create dressed character animations with diverse motions and body shapes, and it can be applied to other garments.

\begin{table*}[t!]
\scriptsize
\center
\caption{Summary of the benchmark datasets for physical simulation task.}
\vspace{-3mm}     
\begin{tabular}{c|c|c|c|c}
  \Xhline{1.0pt}
   Dataset name & \makecell{Publish\\time} & \makecell{\# of\\ photos} & Key features & Sources\\
   \hline
    \makecell[l]{{DeepWrinkles}~\cite{Zorah2018DeepWrinkles}} & 2018 & 9,213 & \makecell[l]{ Each image contains a colored mesh with 200K vertices.} & \makecell[l]{N/A}\\
  \hline 
    \makecell[l]{Santesteban~\emph{et al.}~\cite{Santesteban2019cgf}} & 2019 & 7,117 & \makecell[l]{It was simulated with 17 body shapes with only one garment.} & \makecell[l]{SMPL}\\
  \hline 
    \makecell[l]{\href{https://virtualhumans.mpi-inf.mpg.de/mgn/}{MGN}~\cite{mgn2019iccv}} & 2019 & N/A & \makecell[l]{It contains 356 3D scans of people with different body shapes, poses\\ and in diverse clothing.} & \makecell[l]{SMPL+G}\\
  \hline
    {\makecell[l]{\href{https://renderpeople.com/ 3d- people.}{RenderPeople}}} & 2019 & N/A & \makecell[l]{It consists of 500 high-resolution photogrammetry scans.\\ It was used by PIFu~\cite{pifu2019iccv,pifuHD2020cvpr} and ARCH~\cite{arch2020cvpr}} & \makecell[l]{RenderPeople}\\
  \hline
    {\makecell[l]{\href{https://kv2000.github.io/2020/03/25/deepFashion3DRevisited/}{DeepFashion3D}~\cite{deepfashion3d2020eccv}}} & 2020 & N/A & \makecell[l]{It contains 2,078 3D garment models with 10 different clothing\\ categories and 563 garment instances.} & \makecell[l]{Reconstructed from\\real garments}\\
  \hline
    {\makecell[l]{\href{https://github.com/zycliao/TailorNet_dataset}{TailorNet}~\cite{tailornet2020cvpr}}} & 2020 & 55,800 & \makecell[l]{It contains 20 aligned real static garments with 1,782 different poses\\ and 9 body shapes.} & \makecell[l]{Simulated by the\\ Marvelous Designer }\\
  \hline
    {\makecell[l]{\href{https://virtualhumans.mpi-inf.mpg.de/sizer/}{Sizer}~\cite{sizer2020eccv}}} & 2020 & N/A & \makecell[l]{It includes 100 different subjects with 10 casual clothing classes in\\ various sizes in total of over 2,000 scans.} & \makecell[l]{self-collected}\\
  \Xhline{1.0pt}
\end{tabular}
    \begin{tablenotes}
      \scriptsize
      \item N/A: there is no reported information to cite. M means million.
    \end{tablenotes}
\label{table:physical simulation_dataset}
\end{table*}

\subsubsection{Performance comparison}
\begin{figure}[t!]
  \centering
  \includegraphics[width=\linewidth]{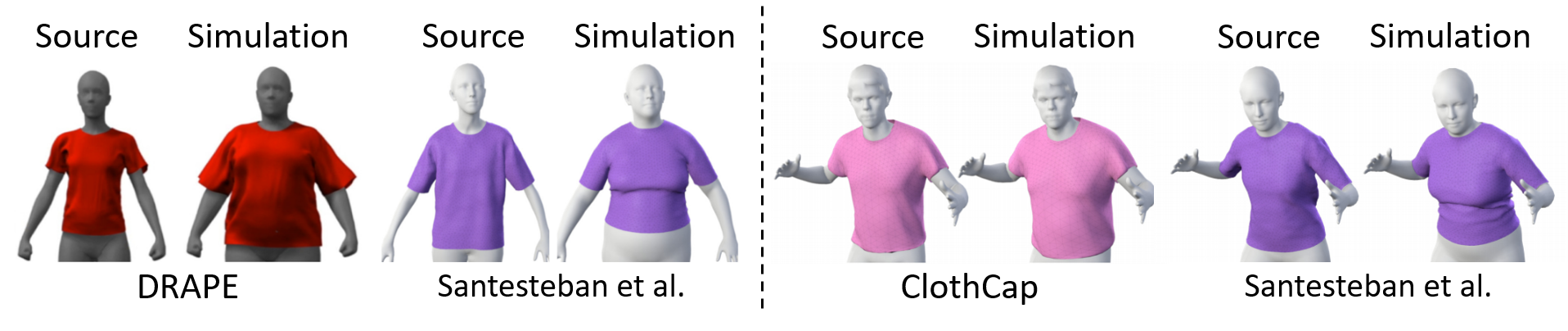}\\
  \vspace{-4mm}
  \caption{Left comparison is between DRAPE~\cite{guan:tog2012} and Santesteban~\emph{et al.}~\cite{Santesteban2019cgf}, while the right one compares between ClothCap~\cite{pons-moll:tog2017} and~\cite{Santesteban2019cgf}. Both are given a source and simulate the physical clothing deformation in different body shapes.}
  \vspace{-5mm}
  \label{fig:physical_simulation_comparsion}
\end{figure}

There are limited quantitative comparisons between physical simulation works. Most of them tend to calculate the qualitative results only within their work (e.g., per-vertex mean error
) or show the vision comparison with state-of-the-art methods. Take the comparison done by
~\cite{Santesteban2019cgf} for example in Fig.~\ref{fig:physical_simulation_comparsion}.

\section{Fashion Recommendation}
\label{sec:fffashion_recommendation}

\textit{``Dressing well is a form of good manners.''} --- Tom Ford \textit{(1961-)}
, while not everyone is a natural-born fashion stylist. In support of this need, fashion recommendation has attracted increasing attention, given its ready applications to online shopping for fashion products. Relevant literature on the research progress of fashion recommendation falls into three main tasks: fashion compatibility, outfit matching, and hairstyle suggestion.


\subsection{Fashion Compatibility}
Fashion recommendation works based on fashion compatibility, which performs how well items of different types can collaborate to form fashionable outfits. Also, it is worth mentioning that the main difference between fashion item retrieval (Sec.~\ref{subsec:item_retrieval}) and fashion recommendation (Sec.~\ref{sec:fffashion_recommendation}) is that the former learns the visual similarity between the same clothing type. In contrast, the latter learns both visual similarity and visual compatibility between different clothing types.

\subsubsection{State-of-the-art methods} 

Veit~\emph{et al.} introduced Conditional Similarity Networks (CSNs) \cite{AVeit2017CVPR}, which learned non-linear feature embeddings that incorporated various notions of similarity within a shared embedding using a shared feature extractor. The CSNs addressed the issue of a standard triplet embedding that treated all triplets equally and ignored the sources of similarity. Song~\emph{et al.}~\cite{song2017neurostylist} integrated visual and contextual modalities of fashion items by employing the autoencoder neural model to seek the non-linear latent compatibility space. Following ~\cite{song2017neurostylist}, Song~\emph{et al.}~\cite{song:sigir2018} integrated fashion domain knowledge to the neural networks to boost the 
performance. Vasileva~\emph{et al.}~\cite{vasileva:eccv2018} presented 
to learn an image embedding that respected item type. They first learned a single, shared embedding space to measure item similarity, then projected from that shared embedding to subspaces identified by type. For learning the compatibility between clothing styles and body shapes, Hidayati~\emph{et al.}~\cite{hidayati:mm2018,hidatayi2020tmm} exploited human body measurements and images of stylish celebrities. They presented a body shape calculator to determine the type of body shape based on a set of body measurements, and a style propagation mechanism to model the correlation between body shapes and clothing styles by constructing graphs of images based on the body shape information and the visual appearance of clothing items, respectively. For using category complementary relations to model compatibility, Yang~\emph{et al.}~\cite{Yang2018aaai} proposed a translation-based neural fashion compatibility model which contained three parts: (1) first mapped each item into a latent space via two CNN for visual and textual modality, (2) encoded the category complementary relations into the latent space, and (3) minimized a margin-based ranking criterion to optimize both item embeddings and relation vectors jointly. 

For making the fashion compatibility task more user-friendly, Wang~\emph{et al.}~\cite{wang2019acmmm} introduced a diagonal process for giving information about which item made the outfit incompatible. They presented an end-to-end multi-layered comparison network to predict the compatibility between different items at different layers and use the backpropagation gradient for diagnosis. Hsiao~\emph{et al.}~\cite{hsiao2019fashionplus} proposed Fashion++ to make minimal adjustments to a full-body clothing outfit that have a maximal impact on its fashionability. Besides, Song~\emph{et al.}~\cite{Song2019GPBPR} took user preferences into account to present a personalized compatibility modeling scheme GP-BPR. It utilized two components, general compatibility modeling and personal preference modeling, for evaluating the item-item and user-item interactions, respectively.

\subsubsection{Benchmark datasets}
The most used source for fashion compatibility datasets is the Polyvore fashion website. It is an online shopping website, where the fashion items contain rich annotations, e.g., clothing color, text description, and multi-view outfit images. We list the benchmark datasets for fashion compatibility in Table~\ref{tab:fashion_compatibility_datasets}. 

\begin{table*}[htbp]
\scriptsize
    \centering
        \caption{Summary of the benchmark datasets for fashion compatibility task.}
        \vspace{-3mm}
    \begin{tabular}{c|c|c|c|c|c|c}
    \Xhline{1.0pt}
      \multicolumn{2}{c|}{Dataset name} & \makecell{Publish \\ time} & \makecell{\# of \\ outfits} & \makecell{\# of item\\ categories} & Key features & Sources \\
    \hline 
        \multicolumn{2}{c|}{\makecell[l]{FashionVC~\cite{song2017neurostylist}}} & 2017 & 20,726 & 2 & \makecell[l]{Annotated with title, category, and description} & \makecell[l]{Polyvore.com} \\
    \hline
        \multicolumn{2}{c|}{\makecell[l]{\href{https://github.com/mvasil/fashion-compatibility}{Vasileva~\emph{et al.}}
        ~\cite{vasileva:eccv2018}}} & 2018 & 68,306 & 19 & \makecell[l]{Annotated with outfit and item ID, fine-grained\\ item type, title, and text descriptions} & \makecell[l]{Polyvore.com} \\
    \hline
        \multirow{6}{*}{\makecell[l]{\href{http://bit.ly/Style4BodyShape}{Style}\\ \href{http://bit.ly/Style4BodyShape}{for} \\\href{http://bit.ly/Style4BodyShape}{Body}\\\href{http://bit.ly/Style4BodyShape}{Shape}
        \\\cite{hidayati:mm2018}}} & \makecell[l]{Stylish Celeb-\\rity Names} & \multirow{6}{*}{2018} & \multirow{3}{*}{N/A} & \multirow{6}{*}{N/A} & \makecell[l]{270 names of the top stylish female celebrities} & \makecell[l]{Ranker.com,\\ fashion magazine sites}\\
    \cline{2-2}\cline{6-7}
         & \makecell[l]{Body\\Measurements} &  &  & & \makecell[l]{Body measurements of 3,150 female celebrities} & \makecell[l]{Bodymeasurements.org}\\
    \cline{2-2}\cline{4-4}\cline{6-7}
         & \makecell[l]{Stylish Celeb-\\rity Images} &  & 347,948 & & \makecell[l]{Annotated with clothing categories and celebrity\\ names} & \makecell[l]{Google Image Search}\\
    \hline
        \multicolumn{2}{c|}{\makecell[l]{\href{https://github.com/WangXin93/fashion_compatibility_mcn}{PolyVore-T}
        ~\cite{wang2019acmmm}}} & 2019 & 19,835 & 5 & \makecell[l]{Categories includes top, bottom, shoes, bag,\\ and accessory} & \makecell[l]{Dataset collected by\\ \cite{han2017learning} from Polyvore} \\
        
    \hline
        \multicolumn{2}{c|}{\makecell[l]{IQON3000~\cite{Song2019GPBPR}}} & 2019 & 308,747 & 6 & \makecell[l]{Categories contains top, bottom, shoes, accesso-\\ry, dress and tunic, and coat. The outfits within\\ this dataset was created by 3,568 users} & \makecell[l]{The fashion web\\ service IQON} \\
    \Xhline{1.0pt}
    \end{tabular}
    \label{tab:fashion_compatibility_datasets}
    \begin{tablenotes}
      \scriptsize
      \item N/A: there is no reported information to cite.
    \end{tablenotes}
\end{table*}

\subsubsection{Performance evaluations}
\label{sec:fashion_compatibility_PE}
For measuring the performance of fashion compatibility works, area under the receiver operating characteristic curve (AUC) is the most used metric. AUC measures the probability that the evaluated work would recommend higher compatibility for positive set than negative set. The AUC scores range between 0 and 1.

\subsection{Outfit Matching}
\begin{figure}[t!]
  \centering
  \includegraphics[width=\linewidth]{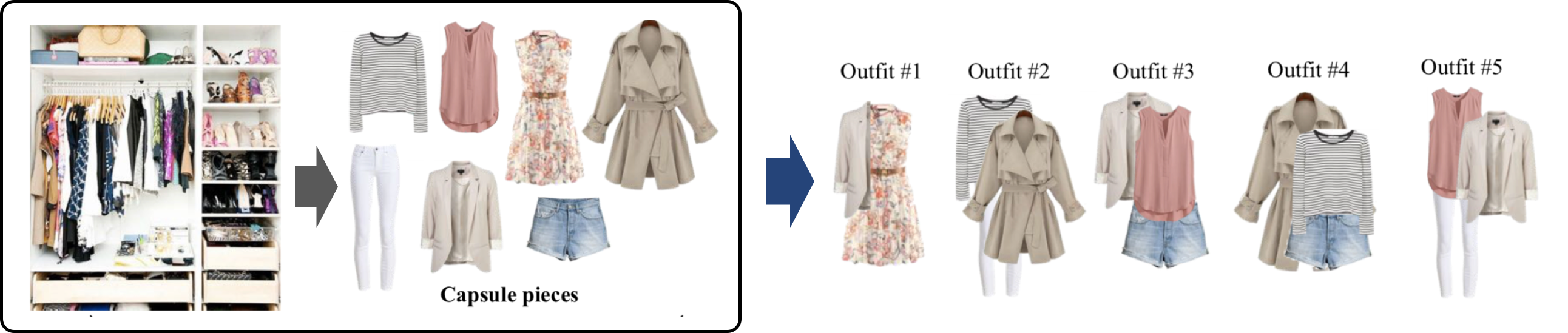}
  \caption{Examples of outfit matching task~\cite{hsiao2018creating}.}
  \label{fig:outfit_matching_example}
\end{figure}

Each outfit generally involves multiple complementary items, such as tops, bottoms, shoes, and accessories. A key to a stylish outfit lies in the matching fashion items, as illustrated in Fig.~\ref{fig:outfit_matching_example}. However, generating harmonious fashion matching is challenging due to three main reasons. First, 
the fashion concept is 
subtle and subjective. Second, there are a large number of attributes for describing fashion. Third, the notion of fashion item compatibility generally goes across categories and involves complex 
relationships. In the past several years, this problem has attracted a great deal of interest, resulting in a long list of algorithms and techniques. 

\subsubsection{State-of-the-art methods}

\begin{figure}[t!]
  \centering
  \includegraphics[width=\linewidth]{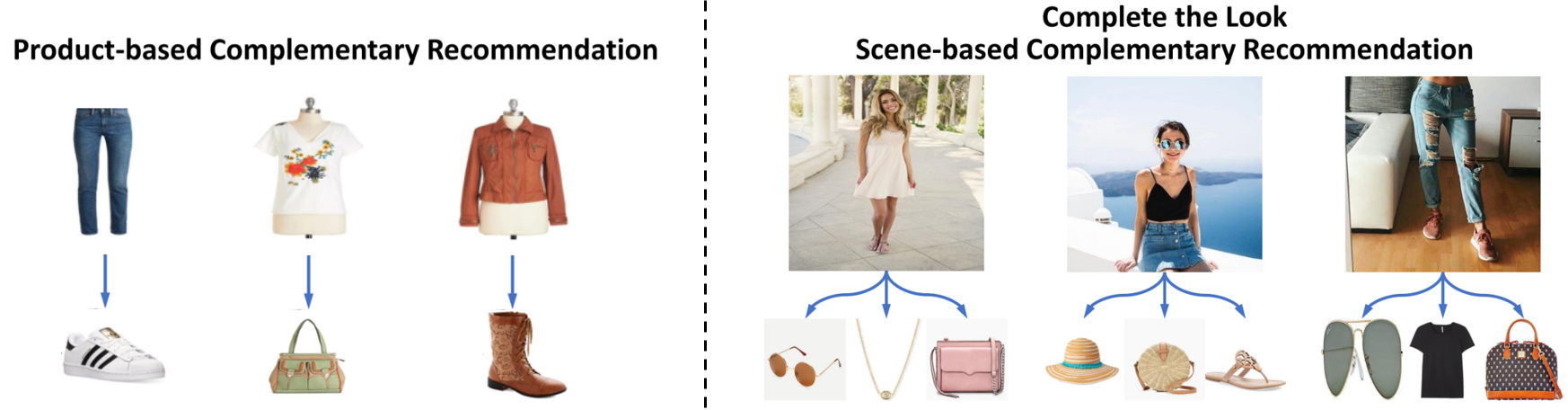}
  \caption{A comparison between product-based and scene-based complementary recommendation~\cite{kang:cvpr2019}.}
  \label{fig:outfit_matching_completeTheLook}
\end{figure}

Fashion recommendation for outfit matching was initially introduced by Iwata~\emph{et al.}~\cite{iwata:ijcai2011}. They proposed a probabilistic topic model for learning information about fashion coordinates. Liu~\emph{et al.}~\cite{liu:mm2012} explored occasion-oriented clothing recommendation by considering the \textit{wearing properly} and \textit{wearing aesthetically} principles. They adopted a unified latent Support Vector Machine (SVM) to learn the recommendation model that incorporates clothing matching rules among visual features, attributes, and occasions. A similar idea of location-oriented recommendation system was proposed by Zhang~\emph{et al.}~\cite{zhang:tmm2017}
. They considered the visual match between the foreground clothing and the background scenery 
and proposed a hybrid multi-label convolutional neural network combined with the 
SVM (mCNN-SVM), which captured the uneven distribution of clothing attributes and explicitly formulated the correlations between clothing attributes and location attributes. More recently, Kang~\emph{et al.}~\cite{kang:cvpr2019} introduced ``Complete the Look'' 
aiming at recommending fashion items that 
go well with 
the given scene. 
They measured both global compatibility (\textit{i.e.}, the compatibility between the scene and product images) and local compatibility (\textit{i.e.}, the compatibility between every scene patch and product image) via Siamese networks and category-guided attention mechanisms. The comparison of the  
product-based and the 
scene-based complementary recommendation is shown in Fig.~\ref{fig:outfit_matching_completeTheLook}. 

A line with metric-based works then proposed to model item-to-item compatibility based on co-purchase behavior. Veit~\emph{et al.}~\cite{veit:iccv2015} utilized the co-purchase data from Amazon.com to train a Siamese 
CNN 
to learn style compatibility across categories and used a robust nearest neighbor retrieval to generate compatible items. The study in~\cite{mcauley:sigir2015} modeled human preference to uncover the relationships between the appearances of pairs of items by Low-rank Mahalanobis Transform that mapped compatible items to embeddings close in the latent space. He and McAuley~\cite{he:aaai2016} later extended the work of~\cite{mcauley:sigir2015} by combining visual and historical user feedback data. The proposed study incorporated visual signals into Bayesian Personalized Ranking with Matrix Factorization as the underlying predictor. 
Besides, 
Hu~\emph{et al.}~\cite{hu:mm2015} addressed the personalized issue by utilizing a functional tensor factorization method to model the user-item and item-item interactions. 


Most previous works mainly focused on top-bottom matching. However, an outfit generally includes more items, such as shoes and bags
. To address this issue, Chen and He~\cite{chen2018dress} extended the traditional triplet neural network, which usually receives three instances, to accept multiple instances. A mixed-category metric learning method was proposed to adapt the multiple inputs. A similar idea was also employed in~\cite{shih2018compatibility}, where the 
proposed generator, referred to as metric-regularized cGAN, was regularized by a projected compatibility distance 
function. It ensured the compatible items were closer in the learned space compared to the incompatible ones.



As for methods with the end-to-end framework, Li \emph{et al.}~\cite{YLi2017TMM} formulated the problem as a classification task, where a given outfit composition was labeled as a popular or unpopular one. They designed a multi-modal multi-instance model, that exploited images and meta-data of fashion items, and information across fashion items, to evaluate instance aesthetics and set compatibility simultaneously. 
Inspired by image captioning of~\cite{donahue2015long}, Han~\emph{et al.}~\cite{han2017learning} built a model based on bidirectional LSTM (Bi-LSTM) to treat an outfit as a sequence of fashion items and each item in the outfit as a time step. A Bi-LSTM model was then utilized to predict the next item conditioned on previously seen ones, where the objective was to maximize the total probability of the given positive sequence. The model was able to achieve three tasks, \emph{i.e.}, fill-in-the-blank (fashion item recommendation given an existing set), outfit generation conditioned on users' text/image inputs, and outfit compatibility prediction. Another work~\cite{hsiao2018creating} also borrowed the idea from natural language processing, which meant that an outfit was regarded as a ``document'', an inferred clothing attribute was taken to be a ``word'', and a clothing style was referred to the ``topic''. The problem of outfit matching was formulated with the topic model. The combinations similar to previously assembled outfits should have a higher probability, which can be employed as the prediction for compatibility, and then solve the outfit matching problem.

For building the bridge between fashion compatibility and personalized preference in outfit matching tasks, there are a few methods for this goal. 
A personalized clothing recommendation system, namely i-Stylist that retrieved clothing items through the analysis of user's images, was developed in \cite{sanchez-riera:mmm2017}. The i-Stylist organized the deep learning features and clothing properties of user's clothing items as a fully connected graph. The user's personalized graph model later derived the probability distribution of the likability of an item in shopping websites. Dong~\emph{et al.}~\cite{dong2019acmmm} took user preference and body shape into account for measuring the user-garment compatibility to deal with personalized capsule wardrobe creation task. They introduced an optimization-based  
framework with dual compatibility modeling, which can both evaluate the garment-garment compatibility and user-garment compatibility. Besides, Yu~\emph{et al.}~\cite{Yu2019ICCV} worked for synthesizing new items automatically for recommendation. Given a query item, the personalized fashion design network they proposed would generate a fashion item for the specific user based on fashion compatibility and user preference. Furthermore, Chen~\emph{et al.}~\cite{chen2019kdd} presented an industrial-scale Personalized Outfit Generation (POG) model. 
They deployed POG on platform \textit{Dida} in \textit{Alibaba} to recommend personalized fashionable outfits for users. For providing more concrete recommendation for users, Hou~\emph{et al.}~\cite{Hou2019ijcai} proposed a semantic attribute explainable recommender system to not only recommend for personalized compatible items but also explain the reason why the system recommends it.

\subsubsection{Benchmark datasets}

Since different papers are under various settings and most of the datasets for outfit matching are not publicly available, almost every 
work collected their own dataset. 
We list the benchmark datasets for outfit matching in Table~\ref{tab:outfit_matching_datasets}. Note that the outfit database in~\cite{mcauley:sigir2015, he:aaai2016, hu:mm2015, YLi2017TMM, han2017learning, hsiao2018creating, kang:cvpr2019} 
comprises the images of each single item, while in~\cite{iwata:ijcai2011, liu:mm2012, chen2018dress} consists of an outfit in a single image.

\begin{table*}[t!]
\scriptsize
    \centering
        \caption{Summary of the benchmark datasets for outfit matching task.}
        \vspace{-3mm}
    \begin{tabular}{c|c|c|c|c|c}
    \Xhline{1.0pt}
      \makecell{Dataset name} & \makecell{Publish \\ time} & \makecell{\# of \\ outfits} & \makecell{\# of item\\categories} & Key features & Sources \\
    \hline
        \makecell[l]{What-to-Wear\\\cite{liu:mm2012}} & 2012 & 24,417 & 2 &  \makecell[l]{Occasion-oriented work; Annotated with full-body,\\ upper-body, or lower-body} & \makecell[l]{Online shopping\\ photo sharing sites}\\
    \hline
        \makecell[l]{Styles and\\Substitutes \cite{mcauley:sigir2015}} & 2015 & 773,465 & N/A & \makecell[l]{Annotated with 4 categories of relationship: (1) users\\ who viewed X also viewed Y, (2) users who viewed\\ X eventually bought Y, (3) users who bought X also\\ bought Y, (4) users bought X and Y simultaneously} & \makecell[l]{Amazon.com}\\
    \hline
        \makecell[l]{Hu~\emph{et al.}~\cite{hu:mm2015}} & 2015 & 28,417 & 3 & \makecell[l]{Annotated with categories, names, and descriptions} & \makecell[l]{Polyvore.com}\\
    \hline
        \makecell[l]{He~\emph{et al.}~\cite{he:aaai2016}} & 2016 & 598.353 & N/A & \makecell[l]{Annotated with users' review histories} & \makecell[l]{Amazon.com,\\ Tradesy.com}\\
    \hline
        {\makecell[l]{Journey Outfit\\\cite{zhang:tmm2017}}} & 2017 & 3,392 & N/A & \makecell[l]{Location-oriented work; Annotated with 14 travel\\ destinations} & \makecell[l]{Online travel\\ review sites}\\
    \hline
        \makecell[l]{Li~\emph{et al.}~\cite{YLi2017TMM}} & 2017 & 195,262 & 4 & \makecell[l]{Annotated with\ title, category, and number of likes} & \makecell[l]{Polyvore.com} \\
    \hline 
        \makecell[l]{Han~\emph{et al.}~\cite{han2017learning}} & 2017 & 21,889 & 8 & \makecell[l]{Each item contains a pair -- product image and a corre\\sponding text description} & \makecell[l]{Polyvore.com} \\
    \hline
        \makecell[l]{Hsiao~\emph{et al.}~\cite{hsiao2018creating}} & 2018 & 3,759 & $\geq 2$ & \makecell[l]{Annotated with meta-labels, \emph{e.g.}, season (winter, spring,\\ summer, fall), occasion (work, vacation), and function\\ (date, hike)} & \makecell[l]{Polyvore.com} \\
    \hline
        \makecell[l]{Fashion\\ Collocation\cite{chen2018dress}} & 2018 & 220,000 & 5 & \makecell[l]{Annotated with independent and ready for wearing (off-\\body module) or dependent (on-body module), or a \\bounding box} & \makecell[l]{Chictopia.com,\\ Wear.net,\\ FashionBeans.com} \\
    \hline
        \makecell[l]{\href{https://github.com/kang205/STL-Dataset}{Pinterest's Shop}\\\href{https://github.com/kang205/STL-Dataset}{The Look}
        ~\cite{kang:cvpr2019}} & 2019 & 38,111 & 10 & \makecell[l]{Annotated with categories of shoes, tops, pants, hand-\\bags, coats, sunglasses, shorts, skirts, earrings, necklaces} & \makecell[l]{Pinterest.com}\\
    \hline
        \makecell[l]{\href{https://anonymousresearch.wixsite.com/pcw-dc}{bodyFashion}
        ~\cite{dong2019acmmm}} & 2019 & 75,695 & N/A & \makecell[l]{It contains 11,784 users with his/her latest historical pur-\\chase records in total of 116,532 user-item records} & \makecell[l]{Amazon.com}\\
    \hline
        \makecell[l]{Yu~\emph{et al.}~\cite{Yu2019ICCV}} & 2019 & 208,814 & N/A & \makecell[l]{It contains 797 users with 262 outfits and each outfit with\\ 2 items, i.e., a top and a bottom} & \makecell[l]{Polyvore.com}\\
    \hline 
        \makecell[l]{\href{https://github.com/wenyuer/POG}{POG}
        ~\cite{dong2019acmmm}} & 2019 & 1.01M & 80 & \makecell[l]{It is composed of 583,000 individual items} & \makecell[l]{Taobao.com,\\iFashion}\\
    \Xhline{1.0pt}
    \end{tabular}
    \label{tab:outfit_matching_datasets}
    \begin{tablenotes}
      \scriptsize
      \item N/A: there is no reported information to cite ; M means million.
    \end{tablenotes}
\end{table*}

\subsubsection{Performance evaluations}
As the evaluation protocol for fashion compatibility (Sec.~\ref{sec:fashion_compatibility_PE}), AUC is the most used metric for outfit matching methods. 
While some methods are also evaluated with NDCG, \textit{i.e.}, \cite{liu:mm2012, hu:mm2015, Hou2019ijcai}, and FITB (fill in the blank) accuracy, \textit{i.e.}, \cite{han2017learning, chen2019kdd}. Unfortunately, there is no unified benchmark for outfit matching, both in datasets and evaluation metrics. Therefore, we are unable to give a comparison of different methods.

\subsection{Hairstyle Suggestion}

Hairstyle plays an essential role in physical appearance. People can look completely different with a different hairstyle. The right hairstyle can enhance the best facial features while concealing the flaws, bringing out natural beauty and style. However, choosing the right hairstyle needs a careful selection as not all hairstyles suit all facial features.

\subsubsection{State-of-the-art methods}
In recent years, many papers have been published related to hairstyles focusing on how to model and render hairstyles with computer graphics~\cite{chai:tog2016, paris:tog2004, selle:tog2008, xu:tog2014, fei:tog2017} or on how to segment hair automatically~\cite{yacoob:tpami2006, rousset:icip2008, wang:accv2010, wang:cvpr2012, qin:icme2017}. Only little studies have been devoted to finding the right hairstyle to suit one's face. In the following, we review the literature concerning hairstyle recommendation.

The pioneering work in~\cite{yang:mmm2012} suggested suitable hairstyles for a given face by learning the relationship between facial shapes and successful hairstyle examples. The proposed example-based framework consisted of two steps: the statistical learning step and the composition step. The purpose of the statistical learning step was to find the most suitable hairstyle through Bayesian inference-based model that estimated the probability distributions of hairstyles to a face image. They proposed to use the ratio of line segments as the feature vector for characterizing the shape of each face, and $\alpha$-matting-based method to indicate hair area in the image. The most suitable hairstyle obtained from the statistical learning step was further superimposed over a given face image to output the face in a suitable hairstyle.

Liu~\emph{et al.} later developed the Beauty e-Experts system~\cite{liu:tomm14} to automatically recommend the most suitable facial hairstyle and makeup, and synthesize the visual effects. They proposed to use the extracted facial and clothing features to simultaneously learn multiple tree-structured super-graphs to collaboratively model the underlying relationships among the high-level beauty attributes (\emph{e.g.}, hair length, and eye shadow color), mid-level beauty-related attributes (\emph{e.g.}, eye shape and mouth width), and low-level image features. Besides, they also proposed a facial image synthesis module to synthesize the beauty attributes recommended by the multiple tree-structured super-graphs model. 



\subsubsection{Benchmark datasets}

Table~\ref{tab:hairstyle_suggestion_dataset} provides benchmark datasets for assessing the performance of hairstyle suggestion methods. 
It is worth mentioning that Hairstyle30k~\cite{yin2017acmmm} is by far the largest dataset for hairstyle related problems thought the proposed method is not for hairstyle suggestion.


\begin{table*}[t!]
\scriptsize
    \centering
        \caption{Summary of the benchmark datasets for hairstyle suggestion task.}
        \vspace{-3mm}
    \begin{tabular}{c|c|c|c|c}
    \Xhline{1.0pt}
      Dataset name & \makecell{Publish \\ time} & \makecell{\# of \\ photos} &  Key features & Sources \\
    \hline
        \makecell[l]{Yang~\emph{et al.}~\cite{yang:mmm2012}} & 2012 & 84 & \makecell[l]{Frontal face photos} & \makecell[l]{Hair stylists from 3 salons}\\
     \hline
        \makecell[l]{Beauty e-Experts\\\cite{liu:mm2014}} & 2013 & 1,605 & \makecell[l]{It consists of 20 hair color classes, 3 different hair\\ length attributes (long, medium, or short), 3 differ-\\ent hair shape attributes (straight, curled, or wavy),\\ and 2 kinds of hair volume (dense or normal)} & \makecell[l]{Professional hairstyle and makeup\\ websites (\emph{e.g.,} Stylebistro.com)}\\
    \hline
        \makecell[l]{Hairstyle30k\\\cite{yin2017acmmm}} & 2017 & 30,000 & \makecell[l]{It contains 64 different hairstyles} & \makecell[l]{Various web search engines (\emph{e.g.,}\\ Google, Flicker, and Bing)}\\
    \Xhline{1.0pt}
    \end{tabular}
    \label{tab:hairstyle_suggestion_dataset}
\end{table*}

\subsubsection{Performance evaluations}

Yang~\emph{et al.}~\cite{yang:mmm2012} conducted a user study to evaluate the effectiveness of their proposed system, while Liu~\emph{et al.}~\cite{liu:tomm14} computed the NDCG that measures how close the ranking of the top-\textit{k} recommended styles is to the optimal ranking. However, we are unable to give comparisons on different hairstyle suggestion methods due to inconsistent benchmarks for different papers.

\section{Applications and Future Work}
\label{sec:application}

The future of the fashion world will be shaped in large part by advancements in the technology, which is currently creeping into the creative domains by starting to mimic human neurons. In the following, we discuss emerging uses of fashion technology in some application areas and future work that is needed to achieve the promised benefits.

\subsection{Applications}

The most popular AI application from the top global apparel industry leaders 
currently implementing AI appears to be AI chatbots, also called smart assistants, which is used to interact with their customers. The common use-cases that are covered are: (1) Responding to customer service inquiries and providing suggestions related to product searches through a social media messaging platform, \textit{e.g.}, ``Dior Insider'', (2) Helping customers navigate products online or in-store to find product(s) that align with their interests, \textit{e.g.}, ``Levi's Virtual Stylist'', ``VF Corporation'', ``Macy's On Call'', and ``Nordstrom'', and (3) Virtual assistant to encourage exercise/behavior adherence, \textit{e.g.}, ``Nike on Demand''.

Moving forward, AI technology will have explosive growth to power fashion industry. In addition to connecting with the customers with the use of AI chatbots, we identify there are four other ways that AI is transforming the future of fashion and beauty, which include: (1) \textbf{Improving product discovery.} Visual search makes it easier for shoppers to find, compare, and purchase products by taking or uploading a photo. One example is Google Lens\footnote{\url{https://lens.google.com}} that allows mobile users to perform searches for similar styles through Google Shopping from the photos they take. (2) \textbf{Tailor recommendation.} In order to keep costs low, brands need to better predict customer preferences by gathering and analyzing purchase behavior, customer profile, as well as customer feedback. Using this data alongside AI and machine learning allows fashion retailers to deliver a personalized selection of clothes to customers. Stitch Fix\footnote{\url{https://www.stitchfix.com}} is one of the most popular AI fashion companies in this category. (3) \textbf{Reducing product return.} Customers have more options to choose from than ever when it comes to making purchases. To gain the loyalty of the consumers, one recent focus of the fashion retailer has been to extend an excellent customer service experience not only at the point of purchase but at the point of return as well. AI technology has the power to better engage customers with the personalized shopping experience that leads them to make more informed and confident purchase decisions, which in turn helps retailers lower return rates. Sephora\footnote{\url{https://www.sephora.com}} is an example of a retailer that has developed this strategy. (4) \textbf{Powering productivity and creativity.} The promise of AI for fashion brands that can marry design creativity with digital innovation has a powerful competitive advantage in the market. The AI technology enables fashion brands to sift through consumer data to gain insights on which product features their customers are likely to prefer. 

\subsection{Break the Limitations of Annotation}

Large-scale data annotation with high quality is indispensable. However, current studies are mainly based on relatively small-scale datasets, which is usually constrained by annotation costs and efforts. Faced with such enormous amounts of fashion and beauty related data, how to generate high coverage and high precision annotations to considerably reduce the cost while preserving quality remains a hot issue. Therefore, more efforts in the development of cost-effective annotations approach on fashion and beauty related data are necessary to address the problem.

\subsection{Overcome the Industry Challenges}

There are still many challenges in adopting fashion and beauty technologies in industry because real-world fashion and beauty are much more complex and strict than in the experiments. The main issue is related to system performance which is still far from human performance in real-world settings. The demand for a more robust system consequently grows with it. Accordingly, it is crucial to continue to pay attention to handling data bias and variations to improve the true-positive rate while maintaining a low false-positive rate. Moreover, with the rising interest in mobile applications, there is a definite need to perform the task in a light but timely fashion. It is thus also beneficial to consider how to optimize the model to achieve higher performance and better computation efficiency.

\section{Conclusion}
\label{sec:conclusion}

With the significant advancement of information technology, research in computer vision (CV) and its applications in fashion have become a hot topic and received a great deal of attention
. Meanwhile, the enormous amount of data generated by social media platforms and e-commerce websites provide an opportunity to explore knowledge 
relevant to support the development of intelligent fashion 
techniques. Arising from the above, there has much CV-based fashion technology been proposed to handle the problems of fashion image detection, analysis, synthesis, recommendation, and its applications. The long-standing semantic gap between computable low-level visual features and high-level intents of customers now seems to be narrowing down. Despite recent progress, investigating and modeling complex real-world problems when developing intelligent fashion solutions remain challenging. Given the enormous profit potential in the ever-growing consumer fashion and beauty industry, the studies on intelligent fashion-related tasks will continue to grow and expand.

\begin{acks}
This work was supported in part by the Fundamental Research Funds for the Central Universities, the National Natural Science Foundation of China under Contract No.61772043, the Ministry of Science and Technology of Taiwan under Grants MOST-109-2223-E-009-002-MY3, MOST-109-2218-E-009-025 and MOST-109-2218-E-002-015.
\end{acks}

\bibliographystyle{ACM-Reference-Format}
\typeout{}
\bibliography{main}

\end{document}


\title{Supplementary Material for:\\
Fashion Meets Computer Vision: A Survey}

\author{Wen-Huang Cheng}
\affiliation{%
  \institution{National Chiao Tung University and
National Chung Hsing University}
}
\email{whcheng@nctu.edu.tw}

\author{Sijie Song}
\affiliation{%
  \institution{Peking University}
}
\email{ssj940920@pku.edu.cn}

\author{Chieh-Yun Chen}
\affiliation{%
  \institution{National Chiao Tung University}
}
\email{sky4568520.ep05@nctu.edu.tw}

\author{Shintami Chusnul Hidayati}
\affiliation{%
  \institution{Institut Teknologi Sepuluh Nopember}
}
\email{shintami@its.ac.id}

\author{Jiaying Liu}
\authornote{Corresponding author}
\affiliation{%
  \institution{Peking University}
}
\email{liujiaying@pku.edu.cn}

\makeatletter
\let\@authorsaddresses\@empty
\makeatother

\renewcommand{\shortauthors}{W.-H. Cheng et al.}

\maketitle

\section{Supplementary Materials}
\label{sec:supplementary_materials}

\begin{table*}[h]
\renewcommand{\arraystretch}{1.1}
\tiny
    \centering
        \caption{Performance comparisons of fashion parsing methods (in \%).}
        \vspace{-3mm}
    \begin{tabular}{l|l|c|c|c|c|c|c|c|c|c}
    \Xhline{1.0pt}
        \multirow{2}{*}{\makecell{Evaluation\\protocol}} & \multirow{2}{*}{\makecell[l]{Dataset}} & \multirow{2}{*}{\makecell{Method}} & \multicolumn{8}{c}{\makecell{Evaluation metric}}\\
     \cline{4-11}   
        & & & \makecell{mIOU} & \makecell[l]{aPA} & \makecell[l]{mAGR} & \makecell[l]{Acc.} & \makecell[l]{Fg. acc.} & \makecell[l]{Avg. prec.} & \makecell[l]{Avg. recall} & \makecell[l]{Avg. F-1}\\
    \hline\hline
        \multirow{4}{*}{\makecell[l]{\cite{yang:cvpr2014, XLiang2016TMM}}} & \multirow{2}{*}{\makecell[l]{SYSU-clothes}} & \makecell[l]{Yamaguchi~\emph{et al.}~\cite{KYamaguchi2012CVPR}} & -- & 85.97 & 51.25 & -- & -- & -- & -- & --\\
    \cline{3-11}
         &  & \makecell[l]{CCP~\cite{yang:cvpr2014, XLiang2016TMM}} & -- & 88.23 & 63.89 & -- & -- & -- & -- & --\\
    \cline{2-11}
         & \multirow{2}{*}{\makecell[l]{Fashionista}} & \makecell[l]{Yamaguchi~\emph{et al.}~\cite{KYamaguchi2012CVPR}} & -- & 89.00 & 64.37 & -- & -- & -- & -- & --\\
    \cline{3-11}
         &  & \makecell[l]{CCP~\cite{yang:cvpr2014, XLiang2016TMM}} & -- & \textbf{90.29} & \textbf{65.52} & -- & -- & -- & -- & --\\
    \hline\hline
        \multirow{6}{*}{\makecell[l]{\cite{XLiang2015ICCV}}} & \multirow{3}{*}{\makecell[l]{ATR}} & 
         \makecell[l]{Yamaguchi~\emph{et al.}~\cite{KYamaguchi2013, yamaguchi:tpami2014}} & -- & -- & -- & 88.96 & 62.18 & 52.75 & 49.43 & 44.76\\
    \cline{3-11}
         &  & \makecell[l]{Liang~\emph{et al.}~\cite{liang:tpami2015}} & -- & -- & -- & 91.11 & 71.04 & 71.69 & 60.25 & 64.38\\
    \cline{3-11}
         &  & \makecell[l]{Co-CNN~\cite{XLiang2015ICCV}} & -- & -- & -- & \textbf{96.02} & \textbf{83.57} & \textbf{84.95} & \textbf{77.66} & \textbf{80.14}\\
    \cline{2-11}
         & \multirow{3}{*}{\makecell[l]{Fashionista}} & 
        \makecell[l]{Yamaguchi~\emph{et al.}~\cite{KYamaguchi2013, yamaguchi:tpami2014}} & -- & -- & -- & 89.98 & 65.66 & 54.87 & 51.16 & 46.80 \\
    \cline{3-11}
         &  & \makecell[l]{Liang~\emph{et al.}~\cite{liang:tpami2015}} & -- & -- & -- & 92.33 & 76.54 & 73.93 & 66.49 & 69.30\\
    \cline{3-11}
         &  & \makecell[l]{Co-CNN~\cite{XLiang2015ICCV}} & -- & -- & -- & \textbf{97.06} & \textbf{89.15} & \textbf{87.83} & \textbf{81.73} & \textbf{83.78}\\
    \hline\hline
        \multirow{12}{*}{\makecell[l]{\cite{3wayparsing2019}}} & \multirow{3}{*}{\makecell[l]{LIP}} & 
        \makecell[l]{MuLA~\cite{nie2018mula}} & 49.30 & \textbf{88.50} & -- & 60.50 & -- & -- & -- & --\\
    \cline{3-11}
         &  & \makecell[l]{CE2P~\cite{CE2P2019}} & 53.10 & 87.37 & -- & 63.20 & -- & -- & -- & --\\
    \cline{3-11}
         &  & \makecell[l]{Wang~\emph{et al.}\cite{3wayparsing2019}} & \textbf{57.74} & 88.03 & -- & \textbf{68.80} & -- & -- & -- & --\\
    \cline{2-11}
         & \multirow{3}{*}{\makecell[l]{PASCAL-\\Person-Part}} & 
         \makecell[l]{MuLA~\cite{nie2018mula}} & 65.10 & -- & -- & -- & -- & -- & -- & --\\
    \cline{3-11}
         &  & \makecell[l]{PGN~\cite{gong:eccv2018}} & 68.40 & -- & -- & -- & -- & -- & -- & --\\
    \cline{3-11}
         &  & \makecell[l]{Wang~\emph{et al.}~\cite{3wayparsing2019}} & \textbf{70.76} & -- & -- & -- & -- & -- & -- & --\\
    \cline{2-11}
         & \multirow{3}{*}{\makecell[l]{ATR}} & 
         \makecell[l]{Co-CNN~\cite{XLiang2015ICCV}} & -- & 96.02 & -- & -- & 83.57 & \textbf{84.95} & 77.66 & 80.14\\
    \cline{3-11}
         &  & \makecell[l]{TGPNet~\cite{TGPNet2018mm}} & -- & \textbf{96.45} & -- & -- & \textbf{87.91} & 83.36 & 80.22 & 81.76\\
    \cline{3-11}
         &  & \makecell[l]{Wang~\emph{et al.}~\cite{3wayparsing2019}} & -- & 96.26 & -- & -- & \textbf{87.91} & 84.62 & \textbf{86.41} & \textbf{85.51}\\
    \cline{2-11}
         & \multirow{3}{*}{\makecell[l]{CFD +\\Fashionista +\\CCP}} & 
        \makecell[l]{Deeplab~\cite{chen2018deeplab}} & -- & 87.68 & -- & -- & 56.08 & 35.35 & 39.00 & 37.09\\
     \cline{3-11}
         &  & \makecell[l]{TGPNet~\cite{TGPNet2018mm}} & -- & 91.25 & -- & -- & 66.37 & 50.71 & 53.18 & 51.92\\
    \cline{3-11}
         &  & \makecell[l]{Wang~\emph{et al.}\cite{3wayparsing2019}} & -- & \textbf{92.20} & -- & -- & \textbf{68.59} & \textbf{56.84} & \textbf{59.47} & \textbf{58.12}\\
    \Xhline{1.0pt}
    \end{tabular}
    \label{tab:fashion_parsing_results}
    \begin{tablenotes}
      \scriptsize
      \item ``--'' represents detailed results are not available.
    \end{tablenotes}
\end{table*}

\begin{table}[t!]
\scriptsize
    \centering
        \caption{Performance comparisons of some clothing retrieval methods by different evaluation protocols.}
        \vspace{-3mm}
    \begin{tabular}{c|c|c|c|c}
    \Xhline{1.0pt}
        \makecell{Protocol} & \makecell{Evaluation metric} & \makecell{Dataset} & \makecell{Method} & \makecell{Result}\\
    \hline
        \makecell{\cite{ZQCheng2017CVPR}} & \makecell{Top-20 accuracy} & \makecell{Video2Shop} & \makecell{Kiapour~\emph{et al.}~\cite{kiapour:iccv2015} / Wang~\emph{et al.}~\cite{wang:icmr2016} / AsymNet~\cite{ZQCheng2017CVPR}} & \makecell{23.47 / 28.73 / \textbf{36.63}}\\
    \hline
        \makecell{\cite{jiang:mm2016}} & \makecell{Top-20 accuracy} & \makecell{Exact Street2Shop} & \makecell{Jiang~\emph{et al.}~\cite{jiang:mm2016} / Kiapour~\emph{et al.}~\cite{kiapour:iccv2015}} & \makecell{20.35 / \textbf{30.59}}\\
    \hline
        \multirow{2}{*}{\makecell[l]{\cite{ak:cvpr2018}}} & \multirow{2}{*}{\makecell{Top-30 accuracy}} & \makecell{Deep Fashion} & \makecell{AMNet~\cite{zhao:cvpr2017} / FashionSearchNet~\cite{ak:cvpr2018}} & \makecell{24.60 / \textbf{37.60}} \\
    \cline{3-5}
         &  & \makecell{Shopping 100k} & \makecell{AMNet~\cite{zhao:cvpr2017} / FashionSearchNet~\cite{ak:cvpr2018}} & \makecell {40.60 / \textbf{56.60}} \\

    \hline
        \multirow{2}{*}{\makecell[l]{\cite{liao:mm2018}}} & \multirow{2}{*}{\makecell{Recall@20}} & \makecell{Amazon} & \makecell{AMNet~\cite{zhao:cvpr2017} / EI Tree~\cite{liao:mm2018}} & \makecell{31.20 / \textbf{63.60}} \\
    \cline{3-5}
         &  & \makecell{DARN} & \makecell{AMNet~\cite{zhao:cvpr2017} / EI Tree~\cite{liao:mm2018}} & \makecell {60.50 / \textbf{71.40}} \\
    \Xhline{1.0pt}
    \end{tabular}
    \label{tab:clothing_retrieval_results}
    \begin{tablenotes}
      \scriptsize
      \item The best results are highlighted in bold font.
    \end{tablenotes}
\end{table}

\bibliographystyle{ACM-Reference-Format}
\typeout{}
\bibliography{supplement}